\documentclass[10pt,journal,compsoc]{IEEEtran}

\usepackage[nocompress, noadjust]{cite}

\ifCLASSINFOpdf
\else
\fi

\pdfoutput=1

\usepackage{url}
\usepackage{epsfig}
\usepackage{graphicx}
\usepackage{amsmath}
\usepackage{amssymb}
\usepackage{multirow}
\usepackage{colortbl}
\usepackage{color}
\usepackage{threeparttable}
\usepackage{booktabs}
\usepackage{adjustbox}
\usepackage{caption}
\usepackage{subcaption}
\usepackage[misc]{ifsym}
\usepackage{xcolor,colortbl}
\usepackage{makecell}
\newcommand*{\belowrulesepcolor}[1]{%
  \noalign{%
    \kern-\belowrulesep 
    \begingroup 
      \color{#1}%
      \hrule height\belowrulesep 
    \endgroup 
    \vspace{-0.03mm}
  }%
} 
\newcommand*{\aboverulesepcolor}[1]{%
  \noalign{%
  \vspace{-0.03mm}
    \begingroup 
      \color{#1}%
      \hrule height\aboverulesep 
    \endgroup 
    \kern-\aboverulesep 
  }%
}

\usepackage{pifont}

\usepackage{xspace}
\makeatletter
\DeclareRobustCommand\onedot{\futurelet\@let@token\@onedot}
\def\@onedot{\ifx\@let@token.\else.\null\fi\xspace}

\usepackage[linesnumbered,lined,boxed,commentsnumbered,ruled]{algorithm2e}

\usepackage[pagebackref=false,breaklinks=true,colorlinks,citecolor=blue,urlcolor=blue,linkcolor=blue,bookmarks=false]{hyperref}

\begin{document}

\title{Watch, Remember, Reason: \\Human-View Video Understanding with MLLMs}

\author{ 
        Jiahao~Meng,
        Yue~Tan,
        Qi~Xu,
        Kuan~Gao,
        Weisong~Liu, 
        Yanwei~Li,
        Jason~Li,
        Lingdong~Kong,
        Haochen~Wang,
        Qianyu~Zhou,
        Jiangning~Zhang,
        Guangliang~Cheng,
        Yunhai~Tong, 
        Lu~Qi, 
        Minghsuan~Yang
\IEEEcompsocitemizethanks{
\IEEEcompsocthanksitem J. Meng, Y.~Tan, and Y.~Tong are with School of Intelligence Science and Technology, Peking University.
\IEEEcompsocthanksitem Q.~Xu and L.~Qi are with Wuhan University.
\IEEEcompsocthanksitem K.~Gao and Y.~Li are with Shanghai Jiao Tong University.
\IEEEcompsocthanksitem J.~Li is with Nanyang Technological University.
\IEEEcompsocthanksitem H. Wang and W. Liu are with CASIA.
\IEEEcompsocthanksitem Q. Zhou is with the University of Tokyo.
\IEEEcompsocthanksitem G. Cheng is with the University of Liverpool.
\IEEEcompsocthanksitem J. Zhang is with Zhejiang University.
\IEEEcompsocthanksitem L. Kong is with the National University of Singapore.
\IEEEcompsocthanksitem  M. Yang is with UC Merced.
}
}

\markboth{IEEE TRANSACTIONS ON PATTERN ANALYSIS AND MACHINE INTELLIGENCE}{}
%


\IEEEtitleabstractindextext{
\begin{abstract}

Video understanding is being rapidly transformed by multimodal large language models (MLLMs), as research moves from short clips to long, multimodal, and knowledge-intensive video scenarios.
These scenarios require models to handle sparse evidence, long-range dependencies, multimodal alignment, and reliable inference under limited computational budgets.
This work presents a \emph{human-view} perspective on LLM-based video understanding, organized around three functional abilities: \emph{watching}, \emph{remembering}, and \emph{reasoning}.
Rather than treating video tasks as isolated benchmarks, this view provides a unified structure for analyzing how video MLLMs acquire evidence, preserve context, and produce grounded outputs.
We introduce a formulation that characterizes video understanding systems by their perceptual representations, memory states, reasoning traces, and final predictions.
Based on this formulation, we identify challenges in spatio-temporal perception, efficient long-video processing, memory modeling, streaming understanding, and faithful reasoning.
Representative methods are organized by their roles in video MLLM systems.
\emph{Watching} covers fine-grained, comprehensive, audio-visual, and efficient perception.
\emph{Remembering} includes offline and streaming memory, while \emph{reasoning} covers text-only reasoning and thinking with videos.
We further examine application domains such as egocentric, sports, instructional, medical, and narrative videos, and cover training datasets and evaluation benchmarks across task types, supervision formats, modalities, and capability dimensions.
Finally, we outline open problems and future directions for scalable, memory-aware, and evidence-grounded video intelligence.
Related works will be continuously traced at \url{https://github.com/marinero4972/Awesome-HumanView-VideoUnderstanding}.
\end{abstract}

\begin{IEEEkeywords}
Video Understanding, Video Reasoning, Video MLLMs
\end{IEEEkeywords}

}

\maketitle

\IEEEdisplaynontitleabstractindextext

%
\IEEEpeerreviewmaketitle

\section{Introduction}
\label{sec:intro}

\begin{figure*}[t]
    \centering
    \includegraphics[width=0.96\textwidth]{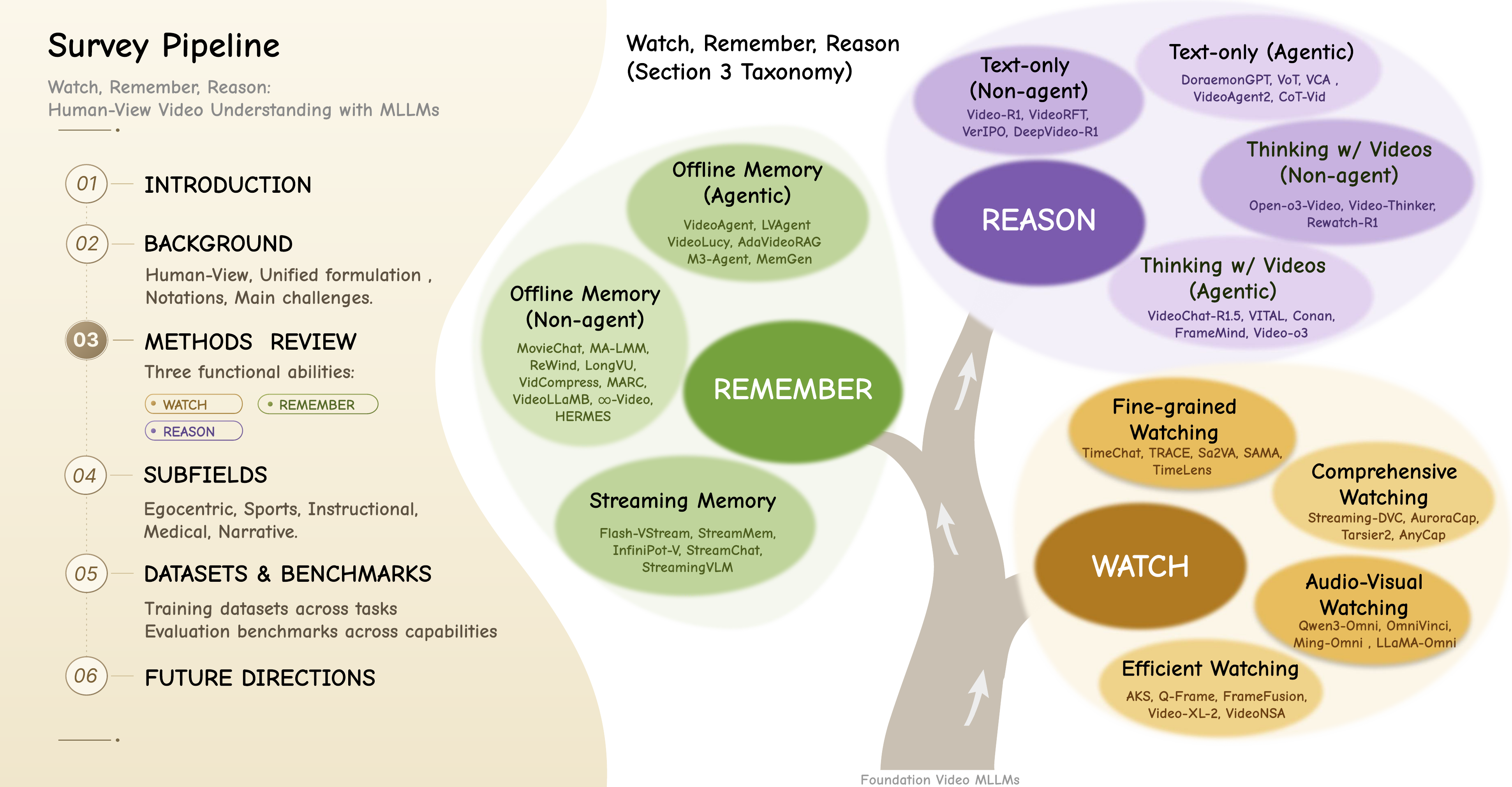}
    \caption{Overview of our survey. \textbf{Left:} the survey pipeline. \textbf{Right:} our \emph{Watch--Remember--Reason} taxonomy for MLLM-based video understanding. \textbf{Watch} (Sec.~\ref{sec:method_watch}) covers fine-grained grounding, captioning, audio-visual perception, and efficient processing. \textbf{Remember} (Sec.~\ref{sec:method_remember}) includes offline and streaming memory. \textbf{Reason} (Sec.~\ref{sec:method_reason}) covers text-only reasoning and thinking with videos, with both agentic and non-agent approaches. Representative methods are listed under each leaf.}
    
    \label{fig:framework}
\end{figure*}

The rapid progress of multimodal large language models (MLLMs) is reshaping video understanding.
Building on large language model (LLM) pre-training, recent multimodal foundation models can process images, videos, and audio~\cite{qwen3.5,qwen3omni,bai2025qwen3vl,xu2025qwen25omni,qin2025videoxl2}.
Early video models mainly focus on short clips and isolated perception tasks~\cite{xu2016msr,xu2017msvdqa,jang2017tgif}.
Recent systems move toward long-horizon comprehension, where inputs may last from minutes to hours~\cite{shen2024longvu,qiu2026longvideo-r1,qin2025videoxl2,zeng2026videoo3}.
This shift changes the core problem.
A model must not only recognize visual content, but also decide what to observe, what to retain, and how to reason over distributed evidence.
These abilities are essential for real-world scenarios such as movies, sports broadcasts, egocentric recordings, instructional lectures, medical procedures, and streaming interactions~\cite{grauman2022ego4d,VideoMMMU,guo2026visually,xu2025streamingvlm}.
In these scenarios, key evidence can be brief, sparse, distant, or scattered across segments.
Therefore, long-form video understanding is not a simple extension of short-video modeling, but requires system designs that jointly consider perception, memory, reasoning, efficiency, and evidence faithfulness.

A central challenge is the tension between redundancy and evidence sparsity.
Long videos contain many redundant frames, yet decisive evidence may appear only briefly.
Thus, models need to selectively perceive useful moments, ground events in time and space, and align visual, audio, and textual signals~\cite{ren2023timechat,tang2025aks,fu2025framefusion}.
They also need compact memory, retrieval, or streaming mechanisms to preserve salient information beyond finite context windows~\cite{song2024moviechat,he2024ma,yuan2025memory,zhang2024flash,yang2025streammem}.
Moreover, many tasks require causal, temporal, spatial, or narrative reasoning over evidence from different moments~\cite{feng2025video-r1,wang2025videorft}.
Faithful reasoning is therefore crucial: models should not only produce plausible answers, but also connect them to explicit spatio-temporal evidence~\cite{meng2025open,zhang2025thinking}.

These challenges suggest studying video understanding as a functional process rather than isolated tasks.
Human video comprehension provides a natural abstraction.
When watching long videos, humans rarely inspect all frames equally; instead, they focus on informative moments, keep useful events in memory, and revisit or connect evidence when answering questions.
This process matches the technical pressures faced by video MLLMs.
Selective observation corresponds to efficient perception and spatio-temporal grounding.
Memory corresponds to retaining long-range context under limited budgets.
Reasoning corresponds to integrating distributed evidence into faithful conclusions.
Therefore, a human-view taxonomy can expose the functional roles of methods and explain why perception, memory, and reasoning need to work together.

Motivated by this view, this survey reviews LLM-based video understanding in terms of three core abilities: \emph{watching}, \emph{remembering}, and \emph{reasoning}.
As shown in Fig.~\ref{fig:framework}, watching focuses on acquiring task-relevant evidence from multimodal video streams.
Remembering focuses on maintaining useful information over long or streaming inputs.
Reasoning focuses on deriving answers from perceived and retained evidence.
This taxonomy connects many recent directions within a single framework, including fine-grained temporal and spatial grounding, long-video efficiency, memory-augmented systems, agentic video understanding, streaming processing, and grounded video reasoning.

Existing surveys have provided valuable summaries of video-language understanding, temporal grounding, post-training, memory, token reduction, and multimodal reasoning~\cite{nguyen2024video,tang2025video-vu-llm,wu2025survey,tang2025video-post,zhang2025survey-rl,hu2025memory,kong2025token,li2025perception}.
For example, Nguyen et al.~\cite{nguyen2024video} review video-language understanding from model architecture, training, and data perspectives, while Wu et al.~\cite{wu2025survey} focus on temporal grounding with MLLMs.
Other surveys cover more specific directions, such as Video-LMM post-training~\cite{tang2025video-post}, memory in AI agents~\cite{hu2025memory}, token reduction~\cite{kong2025token}, and multimodal reasoning models~\cite{li2025perception}.
However, as summarized in Table~\ref{tab:survey-scope}, existing surveys are mostly organized around a specific task, technique, or training paradigm, and thus provide limited integration of perception, memory, and reasoning in long-video MLLM systems.
To fill this gap, our survey makes three main contributions.
\textbf{First,} we introduce a human-view watch--remember--reason taxonomy, which offers a coherent framework for connecting diverse video MLLM methods and clarifying their functional roles.
\textbf{Second,} we provide broad coverage of key techniques in the MLLM era, especially long-video understanding, fine-grained grounding, reasoning, memory, agents, and streaming systems, thereby capturing the frontier of scalable and faithful video intelligence.
\textbf{Third,} we systematically summarize training datasets, evaluation benchmarks, and domain-specific applications, providing practical guidance for model development, evaluation, and future research.

Specifically, we first introduce a unified formulation and notation for video understanding based on the watch--remember--reason process.
This part defines the basic input, output, memory state, and reasoning trace, and clarifies the main challenges faced by video MLLMs.
We then review representative methods from this functional perspective.
First, we examine how models watch videos through temporal and spatial grounding, captioning, omni-modal perception, and efficient visual selection.
Second, we analyze how models remember long contexts through memory compression, hierarchical consolidation, retrieval, and streaming mechanisms.
Third, we discuss how models reason with perceived and retrieved evidence through textual reasoning, agentic tool use, and spatio-temporal grounding.
After that, we discuss several important video subfields, including egocentric, sports, instructional, medical, and narrative videos.
This part shows how different application scenarios impose different requirements on perception, memory, domain knowledge, and reasoning.
We further summarize the training datasets and evaluation benchmarks that support current video MLLMs.
This discussion covers major data types, supervision formats, benchmark dimensions, and evaluation targets.
Finally, we outline open problems and future directions toward scalable, memory-aware, and evidence-grounded video intelligence.

\begin{table*}[t]
\centering
\small
\setlength{\tabcolsep}{4pt}
\scalebox{0.95}{
\begin{tabular}{lccccccccccc}
\toprule
\textbf{Survey} & 
\textbf{TG\&SG} & 
\textbf{Cap} & 
\textbf{Omni} & 
\textbf{Efficiency} & 
\textbf{Off-Mem} & 
\textbf{Streaming-Mem} & 
\textbf{Text-R} & 
\textbf{O3-R} & 
\textbf{Subfields} & 
\textbf{Train-Data} & 
\textbf{Bench} \\
\midrule

\textbf{\cite{tang2025video-post}} 
& $\times$ & $\times$ & $\times$ & $\times$ & $\times$ & $\times$ & $\checkmark$ & $\times$ & $\times$ & $\checkmark$ & $\checkmark$ \\

\cite{tang2025video-vu-llm} 
& $\checkmark$ & $\checkmark$ & $\times$ & $\times$ & $\times$ & $\times$ & $\checkmark$ & $\times$ & $\times$ & $\times$ & $\checkmark$ \\

\cite{zhang2025survey-rl} 
& $\times$ & $\times$ & $\times$ & $\times$ & $\times$ & $\times$ & $\checkmark$ & $\times$ & $\checkmark$ & $\times$ & $\times$ \\

\cite{wu2025survey} & $\checkmark$ & $\checkmark$ & $\times$ & $\times$ & $\times$ & $\times$ & $\times$ & $\times$ & $\times$ & $\times$ & $\checkmark$\\

\cite{nguyen2024video} & $\checkmark$ & $\checkmark$ & $\times$  &  $\times$ & $\times$  & $\times$ & $\times$  & $\times$  & $\times$ & $\checkmark$ & $\times$ \\

\cite{hu2025memory} & $\times$ & $\times$ & $\times$ & $\times$ & $\checkmark$  & $\checkmark$  &  $\times$ &  $\times$ & $\times$ & $\times$ & $\checkmark$ \\

\cite{kong2025token} & $\times$ & $\times$ & $\times$ &  $\checkmark$ &  $\checkmark$ & $\times$ & $\checkmark$  & $\times$  & $\checkmark$ & $\times$  & $\times$ \\

\cite{li2025perception} & $\times$ & $\times$ & $\checkmark$ & $\times$ &  $\checkmark$  & $\times$ & $\checkmark$  & $\times$ & $\times$ & $\checkmark$ & $\checkmark$\\

\textbf{Ours} 
& $\checkmark$ & $\checkmark$ & $\checkmark$ & $\checkmark$ & $\checkmark$ & $\checkmark$ & $\checkmark$ & $\checkmark$ & $\checkmark$ & $\checkmark$ & $\checkmark$ \\

\bottomrule
\end{tabular}
}
\caption{
Comparison of survey scopes under a unified taxonomy.
TG\&SG denotes temporal and spatial grounding.
Cap denotes video captioning.
Omni denotes joint understanding across vision, audio, and language.
Efficiency denotes efficient video processing.
Off-Mem denotes offline memory modeling.
Streaming-Mem denotes online memory mechanisms.
Text-R denotes textual reasoning.
O3-R denotes o3-like video reasoning (thinking-with-videos).
Subfields denotes coverage of domain-specific subfields.
Train-Data denotes coverage of training datasets.
Bench denotes coverage of evaluation benchmarks.
}
\label{tab:survey-scope}
\vspace{-1em}
\end{table*}

\section{Background}

\subsection{Unified View From Human}
\label{sec:background:unified}

Video understanding requires models to process long and complex multimodal streams.
Rather than treating tasks independently, we adopt a unified view based on three core abilities: \textit{watch}, \textit{remember}, and \textit{reason}.
This decomposition follows the human cognitive process and provides a human-centered perspective for understanding diverse video understanding tasks.

\noindent
\textbf{How to Watch.}
Watching corresponds to the perceptual stage of video understanding, where the model selectively attends to visual and auditory signals and forms an initial understanding of what is happening.
It includes identifying when and where events occur, capturing semantic content from scenes, aligning information across modalities, and selecting the most informative evidence under limited computational budgets.

\noindent
\textbf{How to Remember.}
Remembering connects perception with higher-level understanding by retaining salient information over time while discarding redundancy.
It requires the model to preserve both short-term details and long-range context, so that observations from different moments can be accumulated into coherent memory for long-video and streaming scenarios.

\noindent
\textbf{How to Reason.}
Reasoning operates on top of perception and memory to interpret events, infer relations, and produce task-specific outputs.
It may involve multi-step inference over temporally distributed evidence and, in more advanced settings, explicitly ground the reasoning process in visual evidence to improve faithfulness and interpretability.
Overall, this unified view frames video understanding as a progression from perception to memory to reasoning.
It also provides the conceptual basis for the following formulation.

\subsection{Formulation and Notation}
\label{sec:background:notations}

We represent a video as a sequence of frames \(V = \{f_t\}_{t=1}^{N}\), where \(f_t\) denotes the frame at time step \(t\), and \(N\) is the total number of frames.
Additional modalities include audio \(A = \{a_t\}_{t=1}^{N}\), where \(a_t\) is the audio signal at time step \(t\), and optional aligned text \(T = \{\tau_t\}_{t=1}^{N}\), where \(\tau_t\) denotes aligned text such as subtitles, ASR, or captions.

We denote the overall video understanding system by \(\mathcal{F}_{\mathrm{VU}}\).
Given a multimodal video input and a query \(q\), it is defined as \(\mathcal{F}_{\mathrm{VU}} : (V, A, T, q) \rightarrow O\), where \(O\) denotes the output, which may include a textual response, temporal segments, or spatial regions.
Following the watch--remember--reason decomposition, we describe \(\mathcal{F}_{\mathrm{VU}}\) through three functional components: a watching module \(\mathcal{F}_{\mathrm{watch}}\), a memory update module \(\mathcal{F}_{\mathrm{remember}}\), and a reasoning module \(\mathcal{F}_{\mathrm{reason}}\).

\noindent
\textbf{Watching.}
Watching extracts task-relevant perceptual evidence from multimodal video streams.
Since what should be observed often depends on the query, we write
\begin{equation}
Z = \{z_t\}_{t=1}^{N} = \mathcal{F}_{\mathrm{watch}}(V, A, T, q),
\label{eq:watch}
\end{equation}
where \(z_t\) denotes the multimodal representation at time step \(t\).
This stage may include operations such as spatio-temporal grounding, query-aware frame selection, cross-modal alignment, and semantic abstraction.

\noindent
\textbf{Remembering.}
Remembering updates the contextual state over time by accumulating useful evidence and filtering redundancy:
\begin{equation}
m_t = \mathcal{F}_{\mathrm{remember}}(m_{t-1}, z_t, q), \quad t = 1, \dots, N,
\label{eq:remember}
\end{equation}
where \(m_t\) denotes the memory state at time step \(t\), and \(m_0\) is the initial memory.
The memory sequence is denoted by \(M = \{m_t\}_{t=1}^{N}\).

\noindent
\textbf{Reasoning.}
Reasoning operates on perceptual evidence and memory to perform inference:
\begin{equation}
R = \mathcal{F}_{\mathrm{reason}}(Z, M, q),
\label{eq:reason}
\end{equation}
where \(R\) denotes the reasoning trace, which may include textual reasoning steps, grounded evidence such as timestamps and spatial regions, or intermediate tool-use actions.

\noindent
\textbf{Output.}
The final prediction is produced from perception, memory, and reasoning:
\begin{equation}
O = \mathcal{F}_{\mathrm{out}}(Z, M, R, q).
\label{eq:output}
\end{equation}
The above functions are often realized within an MLLM-centered video understanding system.
In practice, such a system may combine the MLLM with external memory, retrieval, or tool modules to support watching, remembering, and reasoning in a unified pipeline.

\noindent
\textbf{MLLM Formulation.}
Let \(x = (V, A, T, q)\) denote the multimodal input, and let \(y = (y_1, \dots, y_L)\) denote the output token sequence.
An autoregressive MLLM parameterized by \(\theta\) defines
\begin{equation}
p_{\theta}(y \mid x)
=
\prod_{i=1}^{L}
p_{\theta}(y_i \mid y_{<i}, x),
\label{eq:mllm}
\end{equation}
where \(y_{<i} = (y_1, \dots, y_{i-1})\).
Under this formulation, the MLLM serves as the core prediction module, while watching, remembering, and reasoning may be implemented through different internal mechanisms or external components within the overall system.
Based on this formulation, modern video understanding systems are typically trained or post-trained under two common paradigms: supervised fine-tuning (SFT) and reinforcement-learning-based post-training such as Group Relative Policy Optimization (GRPO).

\noindent
\textbf{Supervised Fine-Tuning (SFT).}
Given supervised data \(\mathcal{D} = \{(x, y^{*})\}\), SFT optimizes
\begin{equation}
\mathcal{L}_{\mathrm{SFT}}
=
-
\mathbb{E}_{(x, y^{*}) \sim \mathcal{D}}
\left[
\sum_{i=1}^{L}
\log p_{\theta}(y_i^{*} \mid y_{<i}^{*}, x)
\right].
\label{eq:sft}
\end{equation}


\noindent
\textbf{Group Relative Policy Optimization (GRPO).}
For reinforcement-learning-based post-training, GRPO samples a group of outputs \(\{o_i\}_{i=1}^{G}\) for each input \(x\), computes their rewards \(\{R_i\}_{i=1}^{G}\), and normalizes them within the group to obtain relative coefficients \(\tilde{R}_i\).
The objective is
\begin{equation}
\mathcal{L}_{\mathrm{GRPO}}(\theta)
=
-\mathbb{E}
\left[
\frac{1}{G}\sum_{i=1}^{G}
\ell_i(\theta)
\right]
+
\beta D_{\mathrm{KL}}(\pi_{\theta}\|\pi_{\mathrm{ref}}),
\label{eq:grpo}
\end{equation}
where \(G\) is the group size, \(R_i\) is the reward of output \(o_i\), \(\tilde{R}_i\) is the normalized reward within the group, and
\begin{equation}
\ell_i(\theta)
=
\frac{1}{|o_i|}
\sum_{t=1}^{|o_i|}
\min\!\big(
r_{i,t}(\theta)\tilde{R}_i,
\mathrm{clip}(r_{i,t}(\theta),1-\epsilon,1+\epsilon)\tilde{R}_i
\big)
\end{equation}
is the clipped surrogate term.
Here,
\begin{equation}
r_{i,t}(\theta)
=
\frac{
\pi_{\theta}(o_{i,t}\mid x,o_{i,<t})
}{
\pi_{\theta_{\mathrm{old}}}(o_{i,t}\mid x,o_{i,<t})
}
\end{equation}
is the token-level policy ratio, \(\epsilon\) is the clipping coefficient, \(\pi_{\mathrm{ref}}\) is the reference policy, and \(\beta\) controls the strength of the KL regularization.

\subsection{Main Challenges}
\label{sec:background:challenge}

Based on the unified view and formulation above, we summarize the main challenges of video understanding along the three core abilities of watching, remembering, and reasoning.

\noindent
\textbf{Watching.}
Videos are temporally complex: events may be continuous, overlapping, or sparsely distributed, making reliable localization difficult.
At the same time, models must preserve fine-grained spatial details under occlusion, motion blur, and viewpoint changes, while reducing redundancy and aligning asynchronous multimodal signals.

\noindent
\textbf{Remembering.}
Video understanding requires retaining salient information over long durations despite limited context and memory budgets.
Models must selectively preserve important evidence and maintain coherent representations over time, since losing early or subtle cues can lead to incomplete understanding.

\noindent
\textbf{Reasoning.}
Real-world videos often involve complex event structures and long-range dependencies across time and modalities.
Reasoning therefore requires integrating distributed evidence in a stable and scalable manner under constrained computation.

\section{Watch, Remember, Reason: From Functional Perspective}

Based on the unified formulation, we analyze video understanding systems through three core functional abilities: watching, remembering, and reasoning. 
In the following, we review representative methods for each ability, highlighting their design principles, technical variations, and how they address the associated challenges.

\subsection{How to Watch?}
\label{sec:method_watch}

\begin{table*}[t]
\centering
\small
\setlength{\tabcolsep}{6pt}
\renewcommand{\arraystretch}{1.2}
\caption{Representative works about \emph{How to Watch?} (Sec.~\ref{sec:method_watch}).}
\scalebox{0.85}{
\begin{tabular}{p{4.6cm} p{2.2cm} p{2.2cm} p{10.0cm}}
\toprule
\textbf{Method} & \textbf{Year/Conf.} & \textbf{Training} & \textbf{Highlight} \\
\midrule

\multicolumn{4}{l}{\textbf{Section~\ref{sec:watch:fine-grained}: Fine-grained Watching}} \\
\midrule
TimeChat~\cite{ren2023timechat} & CVPR 2024 & SFT & Timestamp-aware encoder with sliding video Q-Former. \\
LITA~\cite{huang2024lita} & ECCV 2024 & SFT & Relative time tokens with SlowFast temporal modeling. \\
UniTime~\cite{li2025unitime} & NeurIPS 2025 & SFT & Interleaved timestamp tokens with adaptive frame scaling. \\
TimeLens~\cite{zhang2025timelens} & CVPR 2026 & SFT+RL & Curated VTG data with RLVR-tuned baseline. \\
OMTG~\cite{xu2026omtg} & ICML 2026 & SFT+RL & One-to-Many Temporal Grounding with RLVR and CoT rewards.\\
Sa2VA~\cite{yuan2025sa2va} & arXiv 2025 & SFT & SAM-2-guided masks in a shared LLM space. \\
SAMA~\cite{sun2025sama} & NeurIPS 2025 & SFT & Context aggregator plus SAM for grounded video chat. \\

\midrule
\multicolumn{4}{l}{\textbf{Section~\ref{sec:watch:captioning}: Comprehensive Watching}} \\
\midrule
Streaming DVC~\cite{zhou2024streaming} & CVPR 2024 & SFT & Fixed-size clustered memory with streaming decoding. \\
DoYouRemember~\cite{doyouremember2024} & CVPR 2024 & SFT & Cross-modal memory retrieval with textual cross-attention. \\
DIBS~\cite{wu2024dibs} & CVPR 2024 & SFT & LLM-generated pseudo boundaries with online refinement. \\
PLLaVA~\cite{xu2024pllava} & arXiv 2024 & Training-free & Parameter-free temporal pooling for dense video captioning. \\
AuroraCap~\cite{auroracap2024} & ICLR 2025 & SFT & Token merging for efficient detailed video captioning. \\
Tarsier2~\cite{yuan2025tarsier2} & arXiv 2025 & SFT+RL & Fine-grained temporal alignment with DPO post-training. \\

\midrule
\multicolumn{4}{l}{\textbf{Section~\ref{sec:watch:omni}: Audio-Visual Watching}} \\
\midrule
Baichuan-Omni~\cite{li2024baichuanomni} & arXiv 2024 & SFT & Progressive multimodal alignment with dedicated video projector. \\
Qwen2.5-Omni~\cite{xu2025qwen25omni} & arXiv 2025 & SFT & TMRoPE for time-interleaved audio-video token alignment. \\
Ming-Omni~\cite{ai2025mingomni} & arXiv 2025 & SFT & Modality-specific MoE routers for unified omni learning. \\
LLaMA-Omni~\cite{fang2024llamaomni} & ICLR 2025 & SFT & Streaming speech decoder for transcription-free voice interaction. \\
Stream-Omni~\cite{zhang2025streamomni} & arXiv 2025 & SFT & Layer-wise speech mapping for simultaneous multimodal interaction. \\
Omni-Captioner~\cite{lu2025omnicaptioner} & ICLR 2026 & SFT & Unified captioner across natural, textual, structured visuals. \\
OmniVinci~\cite{ye2025omnivinci} & ICLR 2026 & SFT & Temporal embedding grouping with constrained rotary time. \\

\midrule
\multicolumn{4}{l}{\textbf{Section~\ref{sec:watch:efficient}: Efficient Watching}} \\
\midrule
AKS~\cite{tang2025aks} & CVPR 2025 & Training-free & Query-relevance and coverage optimized keyframe selection. \\
Q-Frame~\cite{zhang2025qframe} & ICCV 2025 & Training-free & Query-aware frame selection with adaptive multi-resolution scaling. \\
FrameFusion~\cite{fu2025framefusion} & ICCV 2025 & Training-free & Similarity merging plus importance pruning for token reduction. \\
DyCoke~\cite{tao2025dycoke} & CVPR 2025 & Training-free & Temporal merging with dynamic KV cache reduction. \\
Video-XL-2~\cite{qin2025videoxl2} & arXiv 2025 & SFT & Chunk prefilling with bi-level task-aware KV decoding. \\
VideoNSA~\cite{song2025videonsa} & ICLR 2026 & SFT & Hybrid native sparse attention for 128K video contexts. \\
\bottomrule
\end{tabular}
}
\label{tab:watching_papers}
\end{table*}

Watching corresponds to the perceptual stage of video understanding, where models transform raw multimodal inputs into structured representations. 
In this section, we organize existing methods along four complementary dimensions. 
Fine-grained watching focuses on precise spatio-temporal grounding, 
comprehensive watching captures high-level semantic understanding such as captioning and summarization, 
audio-visual watching integrates multimodal signals for coherent perception, 
and efficient watching addresses redundancy and scalability in long videos. 
Together, these dimensions provide a structured view of how video MLLMs perceive and encode visual information prior to memory and reasoning.

\subsubsection{Fine-grained Watching}
\label{sec:watch:fine-grained}

\noindent
\textbf{Temporal Grounding.} 
Video temporal grounding (VTG) aims to localize specific event intervals within untrimmed videos based on natural language queries. 
With the advent of MLLMs, the field is shifting from specialized detection heads toward generative grounding, where timestamps are treated as linguistic tokens within a unified multimodal vocabulary. 
Recent progress can be organized along five axes:
(i)~\emph{time representation}---how timestamps are tokenized and
supervised;
(ii)~\emph{long-video efficiency}---strategies for evidence coverage
under the limited context;
(iii)~\emph{structured decoding}---output formats that reduce temporal
ambiguity;
(iv)~\emph{architecture for fine-grained perception}---encoder designs
that sharpen temporal precision; and
(v)~\emph{verifiable post-training}---reinforcement learning that improves generalization beyond supervised fine-tuning.

\noindent
\textit{Time representation.}
Time-aware instruction tuning makes timestamp prediction a native generation behavior: TimeChat~\cite{ren2023timechat} and VTimeLLM~\cite{huang2024vtimellm} bind visual tokens with timestamps and emphasize boundary-aware recipes.
Later work refines \emph{time tokenization}: LITA~\cite{huang2024lita} uses relative and multi-rate temporal tokens; VTG-LLM~\cite{guo2025vtgllm} injects timestamp information with lightweight compression; DisTime~\cite{zeng2025distime} models time as distributions to better handle ambiguity. 
At the foundation-model level, Qwen3-VL~\cite{bai2025qwen3vl} explicitly upgrades video time modeling via \emph{text--timestamp alignment} and interleaved spatio-temporal positional encoding, making timestamp grounding a built-in capability rather than a task-specific add-on.

\noindent
\textit{Long-video efficiency.}
Evidence coverage is often the bottleneck.
Thus, improving the efficiency of long-video input is also important.
SeViLA~\cite{yu2023sevila} couples query-aware localization with self-chained refinement to reduce dependence on dense labels.
LLaVA-MR~\cite{lu2024llavamr} improves retrieval under context limits via dense time encoding, informative frame selection, and dynamic token compression.
TimeSuite~\cite{zeng2025timesuite} adapts short-video MLLMs to long videos with temporal-aware positional design and grounded tuning data.
For efficiency baselines when temporal search dominates, SOONet~\cite{pan2023soonet} exemplifies scan once end-to-end grounding in long videos.

\noindent
\textit{Structured decoding.}
Structured outputs beyond free-form timestamps reduce underspecification: TRACE~\cite{guo2025trace} generates event-style tuples via causal event modeling.
UniTime~\cite{li2025unitime} combines timestamp-interleaved sequences with coarse-to-fine localization for long videos and multi-event queries.
TAR-TVG~\cite{guo2025tartvg} stabilizes reasoning by inserting multiple timestamp anchors and enforcing anchor-constrained evaluation.

\noindent 
\textit{Architecture for fine-grained perception.} To sharpen temporal precision, models are moving beyond holistic visual encoders. Grounded-VideoLLM~\cite{wang2024grounded} introduces a two-stream encoder that explicitly captures inter-frame relationships via a temporal expert (e.g., InternVideo2) while preserving intra-frame details through a spatial expert encoder. 
Momentor~\cite{qian2024momentor} utilizes a Temporal Perception Module with a continuous interpolation mechanism to address quantization errors in discrete tokens, enabling segment-level reasoning and accurate timestamp prediction. 
VideoPerceiver~\cite{zhao2025videoperceiver} specifically targets transient events (e.g., flicking a switch) by using a key-information-missing training strategy; it replaces key event frames with neighbors and uses an auxiliary contrastive loss to align intermediate representations with motion-sensitive keywords.

\noindent
\textit{Verifiable post-training}
Recent VTG post-training exploits verifiable IoU-style rewards. Time-R1~\cite{wang2025time} shows that RL-style post-training can improve generalization on small, curated data beyond pure SFT.
TimeLens~\cite{zhang2025timelens} argues that a strong and simple default is \emph{timestamp-interleaved} input formatting, and that VTG can benefit from reinforcement learning with verifiable rewards.
OMTG~\cite{xu2026omtg} introduce One-to-Many Temporal Grounding (OMTG), the first task formulation for localizing multiple disjoint segments per query, and establish a comprehensive benchmark, training dataset, and an RLVR pipeline with temporal and caption rewards.
Video-OPD~\cite{li2026videoopd} explores an alternative to GRPO-style RL by on-policy distillation: it samples trajectories from the current policy while using a strong teacher to provide dense token-level supervision, improving training efficiency for TVG. 
%
%
VideoZoomer~\cite{ding2025videozoomer} learns a temporal zoom policy that iteratively requests high-fps clips at selected moments, enabling coarse-to-fine evidence gathering for long-video reasoning under limited frame budgets.
Recipe standardization accelerates adoption: TVG-R1~\cite{chen2025datasets} releases datasets and reproducible RL recipes, while MUSEG~\cite{luo2025museg} targets multi-segment grounding with phased rewards to better handle multiple intervals.


\noindent
\textbf{Spatial-temporal Grounding.} 
This direction mainly comprises two parts: understanding objects using spatio-temporal cues and generating spatio-temporal outputs from given language descriptions, also known as video referring understanding. 
The former means the model needs more fine-grained cues to perform reasoning and obtain answers. 
The latter is the reverse problem, in which the model needs to perform multimodal spatio-temporal perception driven by language. 
We mainly review two directions under the video MLLMs architecture.

\noindent
\textit{Object grounded spatio-temporal understanding.} Existing works that achieve spatio-temporal object understanding can be divided into two directions: tool-use-based methods and naive MLLM architecture with stronger data augmentation. 
Tool-use-based methods use additional models, such as trackers and perception models, to obtain fine-grained visual details. 
For example, VITAL~\cite{zhang2025thinking} proposes a difficulty-aware GRPO and optimizes the model for video-cutting tools. 
As a result, the model can adaptively attend to video tools and integrate their results to form a multimodal CoT.
On the other hand, for MLLMs, recent work also explores stronger data augmentation to achieve spatio-temporal understanding. 
Rex-Omni~\cite{jiang2025detect} introduces a stronger data engine into MLLMs, treating object detection as a point prediction problem and achieving stronger results than MLLM foundation models.
Open-o3-Video~\cite{meng2025open} performs spatio-temporal grounded reasoning within the SFT and RL framework. 
It designs and collects task-specific datasets and uses a box as its reasoning evidence.
STVG-o1~\cite{gu2025thinking} also introduces a bounding-box chain-of-thought mechanism that explicitly reasons about spatio-temporal locations in an intermediate step before producing the final prediction.


\noindent
\textit{Video referring understanding.} Rather than the previous two-stream fusion methods before MLLMs~\cite{UNINEXT}, recent works~\cite{ding2025multimodal} leverage stronger language modeling and image instruction following ability to build more aligned vision-language representation. 
To be more specific, previous image/video referring models~\cite{UNINEXT} adopt DETR-like methods~\cite{zhu2020deformabledetr} to achieve unified segmentation and tracking.
Equipped with LLMs, several recent image referring segmentation and grounding models~\cite{lai2023lisa, OMGLLaVA, qi2024generalizable} have been developed to accomplish more complex referring tasks, including reasoning about referring expressions or joint mask-and-caption generation.
Despite the additional computational requirements of LLMs, the resulting models achieve significant improvements in these referring expression tasks.
One representative method, Sa2VA~\cite{yuan2025sa2va} expands on these studies in the video domain by utilizing SAM-2~\cite{ravi2024sam2}, while maintaining superior performance in both image/video referring tasks and conversation tasks.
Later, several works~\cite{liu2025unipixel} explore better fusion strategies, memory adaptation, and stronger pre-trained MLLMs.
For example, SAMA~\cite{sun2025sama} introduces joint learning of video referring, video grounding, and multi-turn video chat, and presents a simple context aggregator to fuse spatio-temporal features.
More recently, several works~\cite{zhou2026samtok} have explored end-to-end learning with discrete tokenizers to achieve stronger results in LLM-based reinforcement learning.

\begin{figure*}[t]
    \centering
    \includegraphics[width=\textwidth]{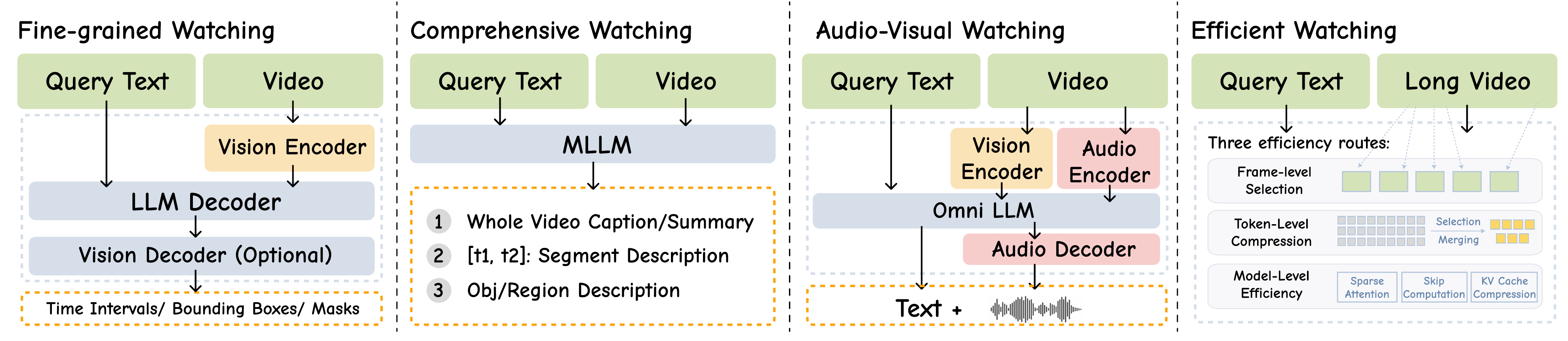}
    
    \caption{Overview of methods related to \textbf{"How to Watch?"}. \textit{Fine-grained watching} localizes task-relevant evidence in time and space. \textit{Comprehensive watching} abstracts videos into summaries, and segment-level or region-level descriptions. \textit{Audio-visual watching} aligns visual and acoustic streams for omni-modal perception. \textit{Efficient watching} reduces redundancy through frame selection, token compression, and efficient model processing.}
    \label{fig:watch}
\end{figure*}

\subsubsection{Comprehensive Watching}
\label{sec:watch:captioning}

Traditional video captioning typically maps a short clip to a single concise sentence,
as in MSVD~\cite{chen2011collecting} and MSR-VTT~\cite{xu2016msr}.
With MLLMs, the task is increasingly framed as open-ended language generation over
visual-token sequences, which makes it natural to describe videos at different
alignment units---the whole video, temporal events, and spatial regions.
We follow this alignment view and discuss whole-video, dense, and region-level
captioning in turn.

%
%
\noindent\textbf{Whole-video Captioning.}
Whole-video captioning aligns language with the entire video, producing either a
concise summary or a detailed description of the main objects, actions, scene
changes, and visual context.
A large portion of recent work adapts image-centric or instruction-tuned LVLMs to
video while controlling the visual-token budget~\cite{maaz2024video,lin2024video,
xu2024pllava,zhang2024llavavideo,auroracap2024}.
Video-ChatGPT and Video-LLaVA show that instruction tuning with unified visual
representations is already enough to obtain strong video-language interaction.
PLLaVA extends an LLaVA-style image model to video via parameter-free adaptive
temporal pooling, handling dense frame inputs without extra temporal
modules~\cite{xu2024pllava}.
AuroraCap keeps the architecture similarly simple but trims redundant visual tokens
through token merging, and comes with VDC, a benchmark for the completeness and
faithfulness of detailed video descriptions~\cite{auroracap2024}.

Going beyond single-pass captions, several methods target hierarchical or
long-context descriptions that connect local actions with global
narratives~\cite{islam2024video,wei2025longcaptioning,chu2025fine}.
Video ReCap recursively generates captions at the clip, segment, and full-video
levels, letting hour-long videos be described from local events up to global
semantics~\cite{islam2024video}.
LongCaptioning focuses on long descriptive outputs and the corresponding benchmark
for long videos, whereas scene-graph consolidation aggregates frame- or
segment-level information into a coherent global description.

Semantic richness and factuality are pushed further by stronger supervision and
post-training objectives~\cite{wang2024tarsier,yuan2025tarsier2,tang2025video,
meng2025videocap,zhong2025owlcap,finegrained2025human}.
Tarsier trains a general fine-grained video description model with large-scale
multi-task supervision, and Tarsier2 tightens frame--event alignment and adds direct
preference optimization for more detailed and accurate
captions~\cite{wang2024tarsier,yuan2025tarsier2}.
Preference- and reward-based approaches build on this idea from different angles:
video-SALMONN 2 optimizes audio-visual caption quality through multi-round DPO,
VideoCap-R1 uses structured reasoning with reward design for action and event
description, and OwlCap relies on reinforcement learning to balance dynamic actions
against static scene details~\cite{tang2025video,meng2025videocap,zhong2025owlcap}.
Human-centric captioning adds structured body priors such as SMPL-based motion
representations to describe pose, body-part interactions, and subtle human motion
more precisely~\cite{finegrained2025human}.

Data construction is the other major driver, with datasets differing in scale,
annotation strategy, and caption style~\cite{chen2024sharegpt4video,chen2024panda,
yang2024vript,zhang2024llavavideo}.
ShareGPT4Video aims at high-quality dense captions via a differential captioning
strategy and then scales annotation with ShareCaptioner-Video~\cite{chen2024sharegpt4video}.
Panda-70M takes the opposite, scale-first route, generating 70M video-caption pairs
with multiple cross-modality teachers and selecting captions over semantically
coherent clips~\cite{chen2024panda}.
Vript provides script-like dense captions for high-resolution videos, and
LLaVA-Video-178K synthesizes video instruction data covering detailed captioning,
open-ended QA, and multiple-choice QA~\cite{yang2024vript,zhang2024llavavideo}.
Together these efforts attack the same bottleneck from complementary
angles---caption quality, corpus scale, script-level detail, and
instruction-following diversity.

Controllable captioning adds user constraints on content, length, format, focus, or
style~\cite{li2025if,anycap2025,qiu2025intentvcnet}.
IF-VidCap evaluates whether models can follow compositional captioning instructions;
AnyCap proposes a plug-and-play residual correction framework that refines captions
from frozen base models toward instruction-compliant
outputs~\cite{li2025if,anycap2025}.
Intent-oriented captioning goes further by conditioning descriptions on user intent
rather than only on visible content~\cite{qiu2025intentvcnet}.

%
%
\noindent\textbf{Dense Video Captioning.}
Dense video captioning aligns language with multiple temporal events inside an
untrimmed video: the output is a set of event captions, each paired with a temporal
interval, coupling event decomposition, timestamp prediction, and sentence generation.
Pre-MLLM work defines this formulation and gradually moves from proposal-based
pipelines to unified generation~\cite{krishna2017dense,zhou2018end,wang2021end,
yang2023vid2seq}.
Krishna et al.~\cite{krishna2017dense} formalize the task with ActivityNet Captions
and a two-stage detect-then-describe pipeline.
Masked Transformer and PDVC replace this with end-to-end transformer decoding and
set prediction over event--caption pairs~\cite{zhou2018end,wang2021end}, and
Vid2Seq represents time boundaries as special tokens in the output sequence,
unifying localization and caption generation in a single sequence-to-sequence
model~\cite{yang2023vid2seq}.

In the video-LLM era, the task becomes tightly coupled with fine-grained temporal
grounding and timestamp-aware generation~\cite{ren2023timechat,huang2024vtimellm,
guo2025vtgllm,guo2025trace,wang2024grounded,wang2024videollm}.
TimeChat embeds timestamps into video encoding and evaluates long-video understanding
jointly through dense captioning, temporal grounding, and highlight
detection~\cite{ren2023timechat}.
VTimeLLM proposes boundary-aware three-stage training to make Video LLMs more
sensitive to temporal boundaries~\cite{huang2024vtimellm}, and VTG-LLM folds
timestamp knowledge into instruction tuning for a family of temporal-grounding tasks
including dense captioning~\cite{guo2025vtgllm}.
TRACE treats the video as a causal event sequence, interleaving timestamps, saliency
scores, and event captions so that later predictions are conditioned on earlier
events~\cite{guo2025trace}.
Grounded-VideoLLM sharpens timestamp prediction with an additional temporal stream
and discrete temporal tokens~\cite{wang2024grounded}, while MMDuet changes the
interface entirely and supports time-sensitive video-text interaction during
continuous playback instead of only producing an offline event
list~\cite{wang2024videollm}.

Memory design becomes central once videos are very long or streamed~\cite{zhou2024streaming,
doyouremember2024,kim2025hicm2}.
Streaming Dense Video Captioning compresses incoming frames into a fixed-size
internal memory and emits event captions causally after events
finish~\cite{zhou2024streaming}.
CM$^2$ explores the complementary, retrieval-augmented route, using cross-modal
memory retrieval to support event localization and caption generation over long
videos~\cite{doyouremember2024}.
HiCM$^2$ extends this with a hierarchical compact memory, suggesting that long-video
DVC benefits from organizing memory at multiple temporal granularities~\cite{kim2025hicm2}.
Annotation is a persistent bottleneck here, since each training video needs both
event boundaries and event descriptions: Vid2Seq partially mitigates this with
large-scale narrated-video pretraining, and DIBS pushes further with pseudo-label
pretraining and online refinement on unlabeled videos~\cite{yang2023vid2seq,
wu2024dibs}.

%
%
\noindent\textbf{Region-level Captioning.}
Region-level video captioning aligns language with spatially localized
targets---objects, regions, masks, or trajectories.
The most common setting starts from a user-specified target and asks the model to
produce a localized description or response~\cite{zhang2025videorefer,pixelrefer2025,
lian2025describe,heo2025omni,tang2025catv,lin2025perceive,qiu2024artemis,
zhou2025strefer}.
VideoRefer packages a full stack for object-level video understanding, with
region-level instruction data, a spatio-temporal object encoder, and a benchmark for
video referring~\cite{zhang2025videorefer}.
PixelRefer improves regional representation through a scale-adaptive object tokenizer
that allocates token resolution according to region size~\cite{pixelrefer2025}, and
Omni-RGPT introduces Token Mark to link visual regions and textual references in a
unified image-video model~\cite{heo2025omni}.
DAM and CAT-V target flexible localized captioning from heterogeneous user inputs:
DAM accepts points, boxes, scribbles, or masks and emits detailed localized captions,
while CAT-V combines segmentation, temporal analysis, and a captioner to describe
user-selected objects over time~\cite{lian2025describe,tang2025catv}.
PAM connects recognition, explanation, captioning, and segmentation for
user-indicated regions in images and videos~\cite{lin2025perceive}.
Artemis tackles cluttered real-world videos via ROI tracking and information-theoretic
target-feature selection~\cite{qiu2024artemis}, and Strefer lowers annotation cost
by synthesizing space-time referring instructions from unlabeled videos~\cite{zhou2025strefer}.

Other work removes the user-specified target and instead discovers object
trajectories automatically~\cite{wong2024elysium,zhou2023dense,fiastre2025maskcaptioner}.
Elysium represents bounding boxes as text tokens so that object tracking and
captioning fit into a single MLLM~\cite{wong2024elysium}.
Dense Video Object Captioning defines a task where models must detect, track, and
caption object trajectories directly from video-level inputs~\cite{zhou2023dense}.
MaskCaptioner goes further and combines open-vocabulary video instance segmentation,
memory-based cross-clip tracking, and trajectory-level caption generation in an
end-to-end model~\cite{fiastre2025maskcaptioner}.
This setting is harder than user-specified description because the model has to
decide which objects are salient, maintain their identities over time, and generate
separate captions for multiple trajectories.

A third thread explicitly grounds phrases or generated text back to video regions,
tying caption generation to spatial localization~\cite{munasinghe2025videoglamm,
yuan2025sa2va,vocap2025,athar2025vicas}.
VideoGLaMM extends grounded video-language interaction to pixel-level
spatio-temporal masks~\cite{munasinghe2025videoglamm}, and Sa2VA pairs SAM2-style
video segmentation with LLaVA-style language modeling for dense grounded
understanding~\cite{yuan2025sa2va}.
VoCap jointly predicts spatio-temporal masklets and object-centric captions from
text, box, or mask prompts~\cite{vocap2025}.
ViCaS complements these with phrase-level annotations that link caption noun phrases
to temporally consistent segmentation masks, so that caption quality and spatial
localization can be evaluated jointly~\cite{athar2025vicas}.

\subsubsection{Audio-Visual Watching}
\label{sec:watch:omni}

The integration of auditory signals---ranging from speech dialogues to environmental sounds---is pivotal for holistic video understanding, as it provides complementary semantic cues often absent in the visual channel. Recent advancements, inspired by the capabilities of GPT-4o~\cite{hurst2024gpt4o}, have driven a paradigm shift from silent video analysis to \textit{Omni-modal} perception, where models are designed not only to watch and listen but also to reason and interact in real time. 
Despite the diversity in applications, contemporary approaches exhibit a homogenized architectural framework comprising modality-specific encoders, projection layers, a unified Large Language Model (LLM) backbone, and modality decoders.

This unified design facilitates the transition from offline perception tasks---such as fine-grained captioning in Omni-Captioner~\cite{lu2025omnicaptioner} and OmniVinci~\cite{ye2025omnivinci}---to dynamic, low-latency interactions.
To enhance reasoning depth within this framework, Omni-R1~\cite{cheng2026omnir1} introduces a two-system collaboration mechanism optimized via Reinforcement Learning (RL).
Simultaneously, architectural efficiency is being actively explored: Ming-Omni~\cite{ai2025mingomni} adopts a Mixture-of-Experts (MoE) strategy with modality-specific routers to balance multi-modal convergence, while Megrez-Omni~\cite{li2025megrezomni} demonstrates that efficient omni-modal perception is achievable even with compact 3B parameters through hardware-software co-design.
Notable interaction-oriented architectures include the ``Thinker-Talker'' design in Qwen2.5-Omni~\cite{xu2025qwen25omni} and Qwen3-Omni~\cite{qwen3omni}, as well as the end-to-end speech interaction capabilities of LLaMA-Omni~\cite{fang2024llamaomni} and InteractiveOmni~\cite{tong2025interactiveomni}. These models extend the scope of video understanding by incorporating streaming generation mechanisms to enable full-duplex human-computer interaction.
A fundamental challenge in this unified framework is the rigorous alignment of asynchronous audio and visual streams within the LLM's context window. To address temporal correspondence, advanced models employ token interleaving strategies combined with explicit timing mechanisms. 
For instance, Qwen2.5-Omni~\cite{xu2025qwen25omni} introduces Temporal Multimodal Rotary Positional Embeddings (TMRoPE) to synchronize audio tokens with their corresponding visual frames, while OmniVinci~\cite{ye2025omnivinci} utilizes OmniAlignNet and temporal embedding grouping to resolve modality misalignment. 
Complementing this temporal synchronization, feature-level alignment is achieved through embedding projection and contrastive learning. Baichuan-Omni~\cite{li2024baichuanomni} mitigates information loss during feature mapping via Conv-GMLP projection. Similarly, Omni-Captioner~\cite{lu2025omnicaptioner} adopts a two-stage training paradigm---freezing the visual encoder to align audio representations---thereby effectively computing an implicit embedding loss akin to CLIP to ensure semantic coherence across modalities.
InteractiveOmni~\cite{tong2025interactiveomni} further strengthens this by employing a cross-modal encoder to fuse features prior to LLM processing, enhancing the model's ability to handle multi-turn dialogue memory.

Beyond input-level synchronization, efficient alignment is critical for the output generation phase, particularly for latency-sensitive streaming applications. 
Connectionist Temporal Classification (CTC) has emerged as a key technique to bridge the gap between semantic understanding and fluid generation. 
LLaMA-Omni~\cite{fang2024llamaomni} leverages a CTC estimator to adaptively align speech representations with textual tokens, facilitating non-autoregressive decoding that significantly reduces latency. 
Building on this, LLaMA-Omni2~\cite{fang2025llamaomni2} further optimizes the decoding process with an autoregressive streaming speech synthesizer, improving the naturalness of the interaction. Similarly, Stream-Omni~\cite{zhang2025streamomni} employs CTC layers to map audio dimensions to text semantics, ensuring that the generated speech remains consistent with the LLM's internal reasoning. By synergizing these alignment strategies---ranging from temporal token interleaving to CTC-based decoding---current research is successfully establishing a robust foundation for real-time, omni-modal video understanding systems.

\subsubsection{Efficient Watching}
\label{sec:watch:efficient}

Efficient perception is vital for understanding long videos.
They often contain redundant information, and only a small part of the content is usually relevant to a specific user question.
Directly feeding all visual tokens into MLLMs often exceeds memory limits and introduces noise that distracts the model.
To address this, recent works~\cite{tang2025aks,zhang2025qframe,fu2025framefusion,qin2025videoxl2,wang2025adaretake} employ selective strategies to reduce input size while preserving key information.
These existing methods can be organized into three categories: frame-level selection, token-level compression and merging, and model-level efficient processing.

\noindent
\textbf{Frame-Level Selection.}
Frame-level strategies filter out irrelevant content before encoding by identifying the most important frames or clips for a given query.
Early representative works mainly focus on query-aware frame filtering.
AKS~\cite{tang2025aks} addresses the trade-off between query relevance and visual coverage through a recursive selection algorithm that maximizes information under a fixed budget.
Q-Frame~\cite{zhang2025qframe} further refines this idea by adjusting the image resolution of selected frames according to their relevance to the user query.
Another line of work introduces more adaptive search mechanisms beyond relevance-only criteria.
Logic-in-Frames~\cite{guo2025logic-in-frames} and DIG~\cite{li2025dig} improve accuracy by verifying visual semantic logic or adapting the search strategy based on the question type.
FOCUS~\cite{lee2025refocus} formulates keyframe selection as a multi-armed bandit problem, enabling efficient discovery of informative regions with minimal exploration.
FrameOracle~\cite{li2025frameoracle} goes one step further by predicting not only frame importance but also the number of frames required to answer a question, trained on 41k annotated examples.
More recent works also move beyond discrete frame selection toward temporally coherent evidence extraction.
F2C~\cite{sun2025frames} and K-Frames~\cite{yao2025k} select continuous clips to better preserve the temporal flow of events, where F2C adopts a training-free strategy, while K-Frames employs a three-stage SFT--RL training pipeline.

\noindent
\textbf{Token-Level Compression and Merging.}
Token-level techniques reduce the number of encoded features by merging similar tokens or removing less important ones.
Representative methods mainly exploit temporal similarity.
FrameFusion~\cite{fu2025framefusion} uses a two-stage strategy that first merges similar tokens in adjacent frames and then removes unimportant ones based on attention scores.
DyCoke~\cite{tao2025dycoke} similarly mitigates temporal redundancy by merging similar tokens across frames based on cosine similarity.
HoliTom~\cite{shao2025holitom} and VidCom$^{2}$~\cite{liu2025vidcom} jointly address inter- and intra-frame compression to enhance the distinctiveness of retained information.
Language-aware methods further make compression adaptive to semantic importance.
LangDC~\cite{wang2025langdc} represents video clips with soft caption tokens produced by a lightweight language model, enabling compression ratios to vary with clip-level semantic density.
DyToK~\cite{li2025dytok} leverages the VLLM's query-conditioned keyframe prior to perform dynamic token allocation, retaining more tokens for salient frames and fewer for redundant ones.

\noindent
\textbf{Model-Level Efficient Processing.}
This category focuses on modifying the model architecture or cache management to handle long video contexts efficiently.
VideoLLM-MoD~\cite{wu2024videollmmod} allows the model to skip computations for redundant tokens at specific layers, thereby reducing unnecessary processing.
AdaRETAKE~\cite{wang2025adaretake} further minimizes information loss by adaptively assigning compression budgets across different model layers and timestamps.
For longer contexts, some methods explicitly optimize the attention or cache mechanism.
Video-XL-2~\cite{qin2025videoxl2} proposes bi-level KV decoding by keeping precise keys and values only for relevant video chunks.
VideoNSA~\cite{song2025videonsa} redesigns the standard attention mechanism with sparse kernels, enabling efficient handling of extremely long sequences at lower cost.

Overall, the field is moving from uniform sampling toward more adaptive and query-aware efficiency mechanisms for long-video understanding.
This trend spans three levels: selecting informative frames, compressing redundant tokens, and optimizing long-context model computation.
Together, these designs reduce computational and memory overhead while preserving task-relevant visual evidence, providing practical support for scaling video MLLMs to hour-long inputs.

\begin{table*}[t]
\centering
\small
\setlength{\tabcolsep}{6pt}
\renewcommand{\arraystretch}{1.2}
\caption{Representative works about \emph{how to remember} (Section~\ref{sec:method_remember})}
\scalebox{0.9}{
\begin{tabular}{p{4.6cm} p{2.2cm} p{2.2cm} p{9.0cm}}
\toprule
\textbf{Model} & \textbf{Year/Conf.} & \textbf{Training} & \textbf{Highlight} \\
\midrule

\multicolumn{4}{l}{\textbf{Section~\ref{sec:remember_offline}: Offline Memory (Agentic)}} \\
\midrule

AVUA~\cite{jeoung2024adaptive} & NeurIPS 2024 & Training-free & Iterative memory use with planning and refinement. \\

VideoAgent~\cite{fan2024videoagent} & ECCV 2024 & Training-free & Tool use centered on temporal and object memory. \\

LVAgent~\cite{chen2025lvagent} & ICCV 2025 & SFT & Uses raw frames directly as memory units. \\

AdaVideoRAG~\cite{xue2025adavideorag} & NeurIPS 2025 & Training-free & Adapts retrieval depth to question difficulty. \\

VideoLucy~\cite{zuo2025videolucy} & NeurIPS 2025 & Training-free & B-tree memory for hierarchical retrieval. \\

GCAgent~\cite{yeo2025gcagent} & arXiv 2025 & Training-free & Event-centric graph memory for structured retrieval. \\

EGAgent~\cite{rege2026agentic} & ACL 2026 & Training-free & Entity-relation graph memory for long videos. \\

MemGen~\cite{zhang2026memgen} & ICLR 2026 & SFT & Token-level memory writing during generation. \\

M3-Agent~\cite{long2026seeing} & ICLR 2026 & SFT + RL & RL-trained retrieval over episodic and semantic memory. \\

\midrule

\multicolumn{4}{l}{\textbf{Section~\ref{sec:remember_offline}: Offline Memory (Non-Agent)}} \\
\midrule

MovieChat~\cite{song2024moviechat} & CVPR 2024 & Training-free & Classic long-short-term memory for video chat. \\

MA-LMM~\cite{he2024ma} & CVPR 2024 & SFT & Autoregressive compression with minimal visual tokens. \\

VidCompress~\cite{lan2024vidcompress} & arXiv 2024 & SFT & Memory-aware temporal compression for long videos. \\

ReWind~\cite{diko2025rewind} & CVPR 2025 & SFT & Read--perceive--write memory preserves chronology. \\

HEM-LLM~\cite{cheng2025enhancing} & ICME 2025 & SFT & Event-level recurrent memory summarization. \\

$\infty$-Video~\cite{santos2025video} & ICML 2025 & Training-free & Continuous-time memory for unbounded videos. \\

LongVU~\cite{shen2024longvu} & ICML 2025 & SFT & Spatiotemporal adaptive compression for long videos. \\

VideoLLaMB~\cite{wang2025videollamb} & ICCV 2025 & SFT & Recurrent memory bridges across long streams. \\

HERMES~\cite{faure2025hermes} & ICCV 2025 & SFT & Episodic and semantic memory for coherent understanding. \\

HierarQ~\cite{azad2025hierarq} & CVPR 2025 & SFT & Hierarchical Q-Former for multi-level memory. \\

MemVid~\cite{yuan2025memory} & arXiv 2025 & SFT + RL & Transformer memory tokens with reasoning cues. \\

MARC~\cite{wu2026marc} & ICLR 2026 & RL & RL-based memory-augmented token compression. \\

\midrule

\multicolumn{4}{l}{\textbf{Section~\ref{sec:remember_streaming}: Streaming Memory}} \\
\midrule

VideoStreaming~\cite{qian2024streaming} & NeurIPS 2024 & Two-stage SFT & Streaming encoding with adaptive memory selection. \\

Flash-VStream~\cite{zhang2024flash} & arXiv 2024 & SFT & Multi-level memory for real-time video streams. \\

StreamChat~\cite{xiong2025streaming} & ICLR 2025 & Training-free & Hierarchical video and dialogue memory for streaming QA. \\

ReKV~\cite{dorovatas2025recurrent} & ICLR 2025 & Training-free & In-context retrieval over historical KV cache. \\

ProVideLLM~\cite{chatterjee2025memory} & ICCV 2025 & Pretrain + SFT & Interleaved visual short-term and textual long-term cache. \\

InfiniPot-V~\cite{kim2025infinipotv} & NeurIPS 2025 & Training-free & Online temporal-spatial KV cache compression. \\

StreamMem~\cite{yang2025streammem} & arXiv 2025 & Training-free & Bounded KV memory via filtering, pruning, and merging. \\

StreamingVLM~\cite{xu2025streamingvlm} & ICLR 2026 & SFT & Constant-memory streaming via sink reuse and windows. \\

\bottomrule
\end{tabular}
}
\label{tab:memory_models}
\end{table*}

\begin{figure*}[t]
    \centering
    \includegraphics[width=0.9\textwidth]{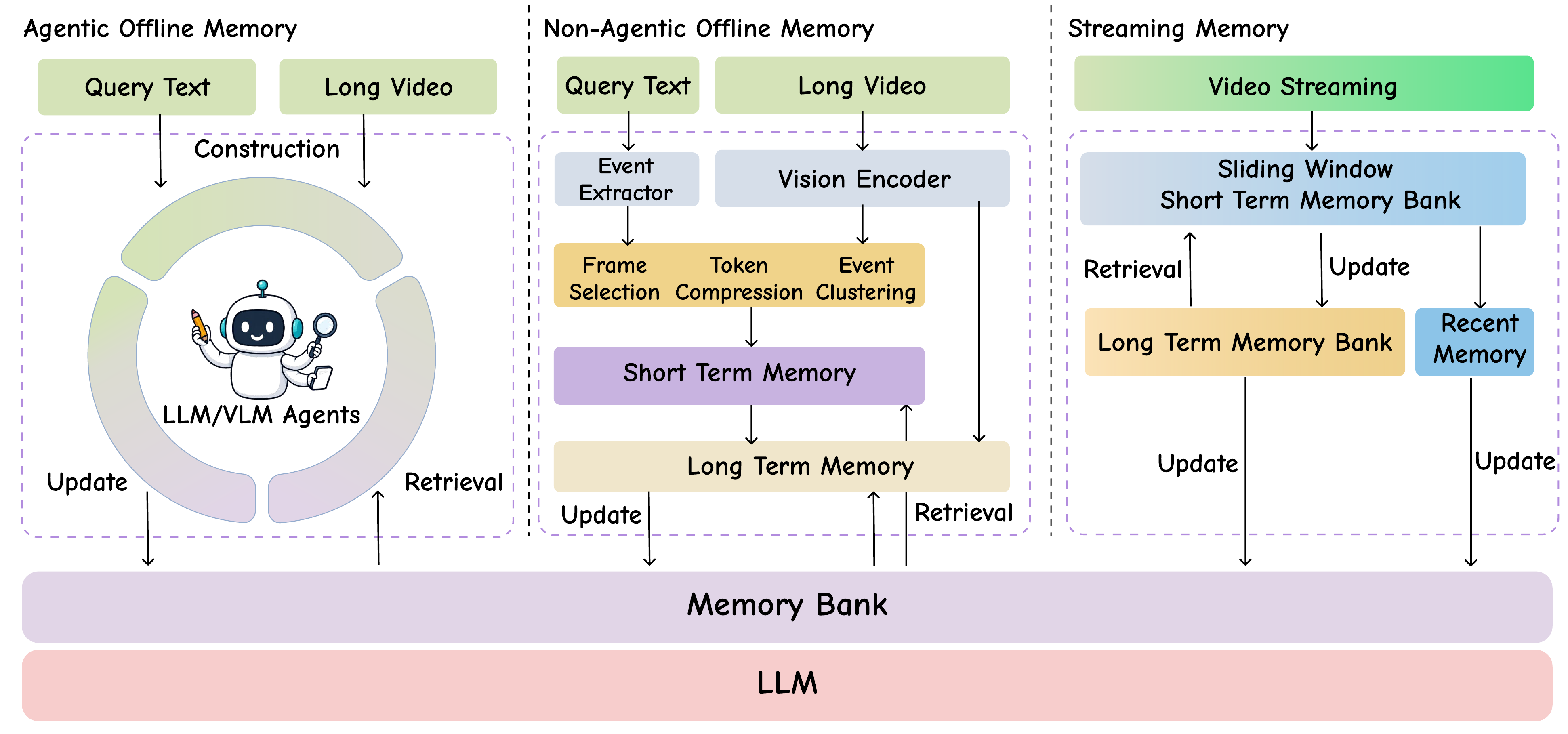}
    \caption{Overview of methods related to \textbf{"How to Remember?"}. \textit{Agentic offline memory} constructs and updates external memory through LLM/VLM agents. \textit{Non-agentic offline memory} builds structured short-term and long-term memory via event extraction, frame selection, token compression, and event clustering. \textit{Streaming memory} maintains and retrieves memory online through sliding windows, recent memory, and long-term memory banks.}
    \label{fig:memory}
\end{figure*}

\subsection{How to Remember}
\label{sec:method_remember}

In long-video understanding, memory enables a model to induce compact representations from extended sequences and to retrieve the appropriate evidence for downstream reasoning. Short-Term Memory (STM) is a transient, task-local state confined to the current context window (e.g., KV caches, clip-level tokens), supporting fine-grained grounding and immediate reasoning. Long-Term Memory (LTM) is persistent information stored externally across tasks and sessions (e.g., vector stores, textual summaries, entity/event graphs), supporting durable recall, aggregation of evidence, and cross-session continuity. Effective systems compress and consolidate STM into LTM to expand capacity while controlling computation overheads.
Streaming Memory is an operational extension of this STM–LTM paradigm that pushes context length toward an effectively unbounded regime. It maintains a rolling STM for recent inputs, performs incremental encoding, selectively writes salient events into LTM, and retrieves from LTM to condition ongoing inference. In doing so, streaming turns finite-window processing into continual, online induction and retrieval under tight latency and resource budgets.

\subsubsection{Offline Memory}
\label{sec:remember_offline}
For exceptionally long or even infinitely long videos~\cite{santos2025video, lin2025unleashing}, the context length limitations of Vision-Language Models (VLMs) necessitate downsampling before the visual sequence can serve as input to answer user queries. The process of storing downsampled information and subsequently extracting desired insights during inference constitutes ``memory'' within the domain of long video understanding. In contrast to Retrieval-Augmented Generation (RAG) techniques---which rely on pre-established external knowledge bases---``memory'' is dynamically generated and written into a memory bank in real time during the model's inference phase.

Typically, a memory system comprises three fundamental modules: memory construction, storage, and retrieval. The core distinction among various memory-based approaches lies in the logical structure of the storage module, with common architectures including long-short-term structures, hierarchical structures~\cite{azad2025hierarq, zuo2025videolucy}, and event-graph structures~\cite{yeo2025gcagent, rege2026agentic}. We adopt a categorization strategy based on the underlying principles of memory construction and utilization, broadly dividing existing methods into two primary paradigms: \emph{Agentic} and \emph{Non-Agent}.

In \textbf{Agentic}
approaches~\cite{fan2024videoagent, jeoung2024adaptive, chen2025lvagent, xue2025adavideorag, yeo2025gcagent, zhang2026memgen, long2026seeing, rege2026agentic, zuo2025videolucy}, Large Language Models (LLMs) or VLMs autonomously invoke memory tools through multi-round reasoning to construct and retrieve memory. While this paradigm benefits from relatively straightforward logical conceptualization and engineering implementation, the multi-turn reasoning process of large models inherently incurs substantial time and computational overhead.

While single MLLMs have improved, they exhibit limitations in modeling long-range dependencies, prompting the adoption of agent-based systems that leverage external tools and multi-agent collaboration. VideoAgent~\cite{fan2024videoagent} established a baseline by adhering to a ``minimal yet effective'' principle, employing an external SQL database to store temporal and object memory. Building on this, recent research focuses on advanced retrieval and planning strategies to improve precision. AdaVideoRAG~\cite{xue2025adavideorag} introduces an adaptive retrieval mechanism that classifies query difficulty to dynamically select retrieval strategies (e.g., graph vs. vector search), while VideoLucy~\cite{zuo2025videolucy} implements a ``deep memory backtracking'' mechanism, akin to a B-tree search, to iteratively refine memory retrieval from coarse to fine granularity. Similarly, MemVid~\cite{yuan2025memory} integrates a learnable memory model to extract reasoning clues before retrieval.

Beyond static retrieval, the field is advancing towards multi-agent collaboration and self-evolution. LVAgent~\cite{chen2025lvagent} and GCAgent~\cite{yeo2025gcagent} propose multi-round dynamic collaboration among MLLM agents to tackle complex queries without task-specific fine-tuning. AVUA~\cite{jeoung2024adaptive} and EGAgent~\cite{rege2026agentic} argue that fixed sampling is suboptimal, instead employing an LLM to propose retrieval strategies and feedback-driven reasoning. Finally, to enable continuous improvement, MemGen~\cite{zhang2026memgen} introduces a generative latent memory using LoRA adapters for self-evolving agents, while M3-Agent~\cite{long2026seeing} combines memorization and control modules to achieve long-term reasoning in real-world environments.

Conversely, \textbf{Non-Agent} methods~\cite{song2024moviechat, he2024ma, lan2024vidcompress, diko2025rewind, cheng2025enhancing, santos2025video, zhang2024flash, shen2024longvu, kim2025infinipotv, xiong2025streaming, wang2025videollamb, faure2025hermes, wu2026marc, yuan2025memory, yang2025streammem, lin2025unleashing, azad2025hierarq, yamao2026question, chen2026learning, jeon2026see} employ a deterministic pipeline for long video understanding, wherein memory construction and retrieval operate as sequentially fixed stages within the pipeline. The primary advantage of this paradigm is that the large model requires only a single forward pass for inference. However, it necessitates additional training or cross-modal alignment, which is prone to introducing extraneous errors.

In non-agent paradigm, the evolution of memory mechanisms has transitioned from dense token retention to sparse, structured consolidation to handle the computational demands of long videos. Early works like MovieChat~\cite{song2024moviechat} pioneered this shift by moving from dense tokens to sparse memory, reducing overhead while maintaining context. This concept was further refined through approaches that focus on dynamic token compression and selection. For instance, ReWind~\cite{diko2025rewind} models memory operations as a Read-Perceiver-Write process to dynamically select frames, while LongVU~\cite{shen2024longvu} and VidCompress~\cite{lan2024vidcompress} employ spatio-temporal adaptive compression to filter redundant visual tokens based on query relevance. To achieve extreme compression, MARC~\cite{wu2026marc} uses Reinforcement Learning (RL) to distill the teacher model's capabilities into a highly compressed 1-frame student memory.

To capture more complex temporal dependencies beyond simple token selection, researchers have shifted towards hierarchical and event-based memory structures. MA-LMM~\cite{he2024ma} and VideoLLaMB~\cite{wang2025videollamb} construct memory banks using sliding windows and recurrent memory bridges, respectively, to maintain continuity. Explicitly decoupling local and global contexts, HierarQ~\cite{azad2025hierarq} and HERMES~\cite{faure2025hermes} utilize hierarchical Q-Formers to separately model episodic and semantic memory. Furthermore, HEM-LLM~\cite{cheng2025enhancing} and its training-free counterpart $\infty$-Video~\cite{santos2025video} propose event-based memory consolidation, clustering adjacent frames into events to support scalable long-term recall, a direction further extended by Hour-LLaVA~\cite{lin2025unleashing} which introduces forgetting mechanisms to handle hour-scale inputs.

\subsubsection{Streaming Memory}
\label{sec:remember_streaming}

Streaming memory addresses the challenge of processing unbounded video streams within fixed memory budgets, necessitating efficient mechanisms to maintain historical context as new inputs arrive. 
A primary strategy is to optimize the Key-Value (KV) cache to prevent memory explosion. StreamMem~\cite{yang2025streammem} and InfiniPot-V~\cite{kim2025infinipotv} propose query-agnostic compression, utilizing attention-based pruning and frame filtering to maintain a constant memory footprint. StreamingTOM~\cite{chen2025streamingtom} introduces Causal Temporal Reduction and Online Quantized Memory to compress tokens before they enter the LLM. Similarly, rLiVS~\cite{dorovatas2025recurrent} and Video-SALMONN S~\cite{sun2025video} employ recurrent selection mechanisms and Test-Time Training (TTT) layers, respectively, to dynamically retain the most relevant historical tokens. StreamingVLM~\cite{xu2025streamingvlm} further achieves constant memory usage via attention sink reuse and sliding windows, enabling infinite stream processing.

To balance recent details with long-term context, many frameworks adopt a dual-memory or hierarchical architecture. Flash-VStream~\cite{zhang2024flash, zhang2025flash} designs a ``Flash Memory'' system where a Frame Handler updates a high-capacity memory while a Question Handler retrieves information, supporting asynchronous parallel processing. StreamChat~\cite{xiong2025streaming} organizes memory into a hierarchical tree structure, separating short-term event tracking from long-term feature compression to support multi-round interaction. ProVideLLM~\cite{chatterjee2025memory} employs an interleaved cache strategy, storing visual tokens in short-term memory and text tokens in long-term memory to maximize compression. StreamForest~\cite{zeng2025streamforest} and VideoStreaming~\cite{qian2024streaming} also contribute to this direction by maintaining persistent event memory and employing memory-propagated streaming encoding, respectively.

Beyond architecture, system-level co-design is crucial for real-time performance. QuickVideo~\cite{schneider2025quickvideo} and LiveVLM~\cite{ning2025livevlm} optimize the prefill and retrieval phases through CPU-GPU parallelism and streaming-oriented KV retrieval, significantly reducing latency. Furthermore, the field is moving towards proactive assistants: StreamBridge~\cite{wang2025streambridge} and Dispider~\cite{qian2025dispider} enable models not only to passively process streams but also to actively perceive, decide, and react to dynamic visual stimuli. 
%

\begin{table*}[t]
\centering
\small
\setlength{\tabcolsep}{6pt}
\renewcommand{\arraystretch}{1.2}
\caption{Representative works about \emph{How to Reason?}(Sec.~\ref{sec:method_reason}).}
\scalebox{0.8}{
\begin{tabular}{p{3.5cm} p{2.3cm} p{2cm} p{13cm}}
\toprule
\textbf{Method} & \textbf{Year/Conf.} & \textbf{Training} & \textbf{Highlight} \\
\midrule
\multicolumn{4}{l}{\textbf{Section~\ref{sec:reason_text}: Text-only Reasoning} \textbf{(Agentic)}} \\
\midrule
VideoAgent~\cite{fan2024videoagent} &  ECCV 2024 & Training-free & Agentic use of temporal/object memory with iterative tool-based reasoning \\
DoraemonGPT~\cite{yang2024doraemongpt} & ICML 2024 & Training-free & Dynamic spatio-temporal memory with Monte-Carlo Tree Search for multi-step explanations \\
Video-of-Thought~\cite{fei2024video} & ICML 2024 & SFT & Video-of-Thought framework for multi-stage perception-to-cognition reasoning \\
VCA~\cite{yang2025vca} & ICCV 2025 & Training-free & Curiosity-driven agent that adaptively explores frames via tree search \\
Flow4Agent~\cite{liu2025flow4agent} & ICCV 2025 & Training-free & Optical-flow motion priors for adaptive temporal granularity and evidence focusing \\
DVD~\cite{zhang2025deep} & NeurIPS 2025 & Training-free & Multi-granularity video database for adaptive agentic search and evidence extraction \\
VideoAgent2~\cite{zhi2025videoagent2} & NeurIPS 2025 & Training-free & Uncertainty-aware retrieval planning to improve multi-step reasoning efficiency \\
CoT-Vid~\cite{jin2025cot} & arxiv 2025 & Training-free & Dynamic CoT routing and self-verification to selectively trigger multi-step video reasoning \\
\midrule
\multicolumn{4}{l}{\textbf{Section~\ref{sec:reason_text}: Text-only Reasoning} \textbf{(Non-agent)}} \\
\midrule
Video-R1~\cite{feng2025video-r1} & NeurIPS 2025 & SFT+RL & Temporal GRPO contrasting ordered vs shuffled frames for robust temporal reasoning \\
TW-GRPO~\cite{dang2025twgrpo} & arXiv 2025 & RL & Token-weighted GRPO emphasizing informative reasoning tokens \\
VistaDPO~\cite{huang2025vistadpo} & ICML 2025 & RL & Hierarchical spatio-temporal DPO to improve text-video alignment and reduce hallucination \\
VerIPO~\cite{li2025veripo} & arXiv 2025 & RL & Verifier-guided iterative policy refinement (GRPO → filter → DPO) \\
VideoRFT~\cite{wang2025videorft} & NeurIPS 2025 & SFT+RL & Semantic-consistency reward guided reinforced fine-tuning for grounded video reasoning\\
Time-R1~\cite{wang2025time} & NeurIPS 2025 & SFT+RL & Temporal IoU + deviation-aware rewards tailored for verifiable temporal grounding \\
DeepVideo-R1~\cite{park2025deepvideo} & NeurIPS 2025 & RL & Difficulty-aware GRPO regression for stable long-video reasoning \\
Video-CoT~\cite{zhang2025video-cot} & ACM MM 2025 & SFT & Spatiotemporal CoT dataset and benchmark for fine-grained video reasoning \\
SpaceR~\cite{ouyang2025spacer} & arXiv 2025 & RL & Spatial map representation with GRPO to enhance spatial reasoning structure \\
\midrule
\multicolumn{4}{l}{\textbf{Section~\ref{sec:reason_o3_like}: Thinking with Videos} \textbf{(Agentic)}} \\
\midrule
VideoChat-R1.5~\cite{yan2025videochat} & NeurIPS 2025 & RL & Temporal or spatial evidence localization via iterative perception \\
Pixel Reasoner~\cite{wang2025pixel} & NeurIPS 2025 & SFT+RL & Pixel-space zoom/crop as explicit reasoning actions, optimized by curiosity-driven RL \\
FrameMind~\cite{ge2025framemind} & arXiv 2025 & RL & Dual-resolution tools for temporal scan and spatial inspection with reinforcement feedback \\
Love-R1~\cite{fu2025love} & arXiv 2025 & SFT+RL & Adaptive slow--fast sampling: dense low-res global scan + clip-level high-res zoom-in \\
VideoZoomer~\cite{ding2025videozoomer} & ICLR 2026 & SFT+RL & Temporal-zoom agent that dynamically controls visual focus to gather evidence \\
VITAL~\cite{zhang2025thinking} & CVPR 2026 & SFT+RL & Temporal retrieval and re-inspection tools for interleaved video reasoning \\
Conan~\cite{ouyang2025conan} & CVPR 2026 & SFT+RL & Multi-scale evidence search and cross-frame detective reasoning \\
VideoTemp-o3~\cite{liu2026videotemp} & ICML 2026 & SFT+RL & Unifies temporal grounding and QA for efficient agentic long-video reasoning \\
Video-o3~\cite{zeng2026videoo3} & ICML 2026 & SFT+RL & Interleaved clue-seeking with decoupled attention for multi-hop evidence search \\
VideoSeek~\cite{lin2026videoseek} & CVPR 2026 & Training-free & Actively seeks sparse answer-critical evidence through multi-granular video logic flow \\
\midrule
\multicolumn{4}{l}{\textbf{Section~\ref{sec:reason_o3_like}: Thinking with Videos} \textbf{(Non-agent)}} \\
\midrule
Video-Thinker~\cite{wang2025video-thinker} & arXiv 2025 & SFT+RL & Video reasoning via structured temporal cues and captions \\
Open-o3-Video~\cite{meng2025open} & ICML 2026 & SFT+RL & Explicit spatio-temporal evidence (timestamps and boxes) in reasoning trace \\
Rewatch-R1~\cite{zhang2025rewatch} & ICLR 2026 & SFT+RL & Re-watching CoT synthesis with observation-consistency reward for grounded reasoning \\
\bottomrule
\end{tabular}
}
\label{tab:reasoning_papers}
\end{table*}

\begin{figure*}[t]
    \centering
    \includegraphics[width=0.9\textwidth]{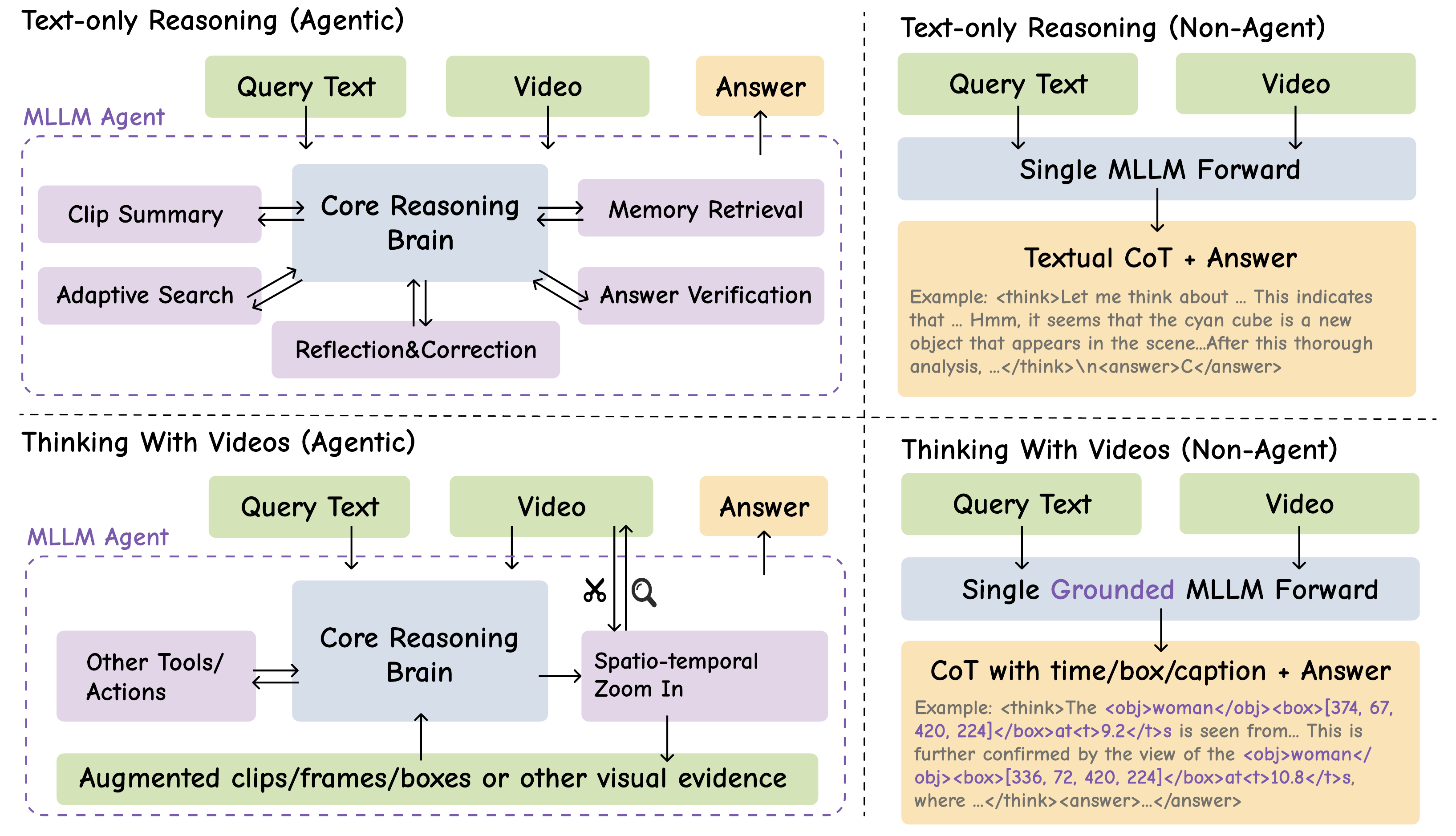}
    
    \caption{Overview of methods related to \textbf{"How to Reason?"}. \textit{Agentic text-only reasoning} methods decompose reasoning into modular steps such as clip summarization, adaptive search, memory retrieval, reflection, and answer verification. \textit{Non-agent text-only reasoning} methods perform a single MLLM forward pass and produces textual chain-of-thought with the final answer. \textit{Agentic thinking with videos} methods actively interact with videos through tools like spatio-temporal zoom-in. \textit{Non-agent thinking with videos} methods directly ground reasoning in visual evidence, such as timestamps, boxes, and captions, within a single grounded MLLM forward pass.}
    
    \label{fig:reason}
\end{figure*}

\subsection{How to Reason?}
\label{sec:method_reason}

A central challenge in video understanding is enabling models to reason over what they have perceived and remembered.
To achieve this, recent works build video reasoning models on top of MLLMs. These models combine semantic recognition, logical inference, and world knowledge to form deeper interpretations of objects, events, and their relations.
They often make their reasoning processes explicit by presenting them as a \texttt{<think>}...\texttt{</think>} trace between two markers, followed by a final answer produced in \texttt{<answer>}...\texttt{</answer>}.
Early studies on video reasoning mainly focus on \textbf{Text-only Reasoning}, where the model reasons purely in language without temporal or spatial grounding.
More recently, many works have moved toward \textbf{Thinking with Videos} (OpenAI-o3-like~\cite{openai-o3}), where the model interleaves reasoning with visual grounding and can actively seek key visual evidence, such as key frames, timestamps, or spatial regions.

\subsubsection{Text-only Reasoning}
\label{sec:reason_text}

Text-only reasoning refers to video understanding methods in which the model explicitly decodes an intermediate thought trace before producing the final answer.
This thinking trace is expressed purely in the language space and may include automatic or prompt-specified components such as video caption and abstraction, step-by-step question analysis, symbolic manipulation, and iterative reflection.
Following the success of DeepSeek-R1~\cite{guo2025deepseek-r1}, text-only video reasoning developed rapidly in video understanding.
Based on how the thinking process is organized, existing methods can be broadly divided into \textbf{agentic approaches} and \textbf{non-agent approaches}.
Agentic approaches treat MLLMs or LLMs as decision-making agents that iteratively plan, retrieve, verify, and revise intermediate results through structured interactions with external tools and memory.
In contrast, non-agent approaches improve reasoning quality mainly through post-training, such as reinforcement learning, preference optimization, or chain-of-thought supervised fine-tuning, without introducing tool-driven control loops.

\noindent
\textbf{Agentic Approaches.}
Agentic approaches construct video reasoning systems by utilizing LLMs or MLLMs in an agent loop~\cite{yuan2025video,zhong2025omni,wang2025video-in-the-loop,jin2025cot,zhang2025vitcot,fan2024videoagent,liu2025videomind,li2025perceive,wang2024videoagent,yang2025vca}.
Through prompting and iterative control, the agent coordinates perception tools, memory modules, and reasoning steps to gradually solve complex queries.
Most agentic methods are training-free, while a small number introduce lightweight supervised or reinforcement learning to specialize certain components~\cite{liu2025videomind,wang2025video-in-the-loop,fei2024video}.

A core challenge in long video reasoning is preserving and organizing information across extended temporal spans.
Following VideoAgent~\cite{fan2024videoagent}, DoraemonGPT~\cite{yang2024doraemongpt} also uses a temporal memory and space memory design, but emphasizes dynamic scene understanding.
It extracts spatio-temporal attributes into queryable memory slots and employs a tool-driven reasoning loop guided by Monte Carlo tree search to explore candidate explanations.
Video-of-Thought~\cite{fei2024video} proposes a step-by-step reasoning framework that explicitly decomposes video understanding from perception to cognition.
Its VoT framework consists of multiple stages, including task definition and target identification, object tracking, action analysis, answer generation, and answer verification, forming a structured reasoning pipeline over intermediate textual representations.
This memory-and-retrieval line is further extended by VideoRAG~\cite{ren2025videorag}, which formulates long video understanding as a retrieval-augmented generation problem.
It builds structured textual representations from ASR and captions, constructs entity-relation graphs across clips, and retrieves relevant segments through both graph-based and embedding-based search before language reasoning.
Related systems such as VideoForest~\cite{meng2025videoforest}, DrVideo~\cite{ma2025drvideo}, and Vgent~\cite{shen2025vgent} further organize video content into hierarchical trees, document-style memories, or semantic graphs, all aiming to reduce the loss of long-range dependencies during multi-step reasoning.

Besides structured memory, another major line of agentic reasoning focuses on adaptive search and selective attention, especially for long videos with heavy redundancy.
For example, VCA~\cite{yang2025vca} introduces a curiosity-driven agent that actively explores long videos by tree search, selecting informative segments until sufficient evidence is collected.
The agent balances exploration and pruning, allowing it to focus on semantically relevant regions while skipping irrelevant content.
DVD~\cite{zhang2025deep} formulates long videos as a multi-granularity database and equips the agent with adaptive search tools, including global browsing, segment retrieval, and frame inspection. The agent autonomously plans tool usage based on intermediate uncertainty, progressively narrowing the search space.
VideoTree~\cite{wang2025videotree} and Flow4Agent~\cite{liu2025flow4agent} also follow related principles by organizing frames into adaptive tree structures or leveraging motion priors from optical flow, enabling agents to allocate attention more densely to action-rich intervals.

More recent methods further strengthen reasoning verification and reflection~\cite{jin2025cot,zhi2025videoagent2,yang2025streamagent,montes2025viqagent}.
CoT-Vid~\cite{jin2025cot} employs dynamic chain-of-thought routing, where the agent first determines whether multi-step reasoning is required and then decomposes the task into summarization, verification, and reflection stages.
Self-consistency and clustering-based validation are used to select reliable reasoning paths and suppress hallucinations.
VideoAgent2~\cite{zhi2025videoagent2} extends earlier agent frameworks by incorporating uncertainty-aware reasoning.
The agent evaluates confidence in preliminary answers and triggers targeted tool-based retrieval only when uncertainty is high, following a plan-and-adjust loop.
Other methods, such as StreamAgent~\cite{yang2025streamagent} and ViQAgent~\cite{montes2025viqagent}, further integrate anticipatory feedback or open-vocabulary grounding validation to refine intermediate conclusions and improve robustness.

Overall, agentic textual reasoning provides a flexible and interpretable paradigm for long video understanding.
By using perception and memory results and introducing iterative control, these methods scale to long-form videos while avoiding heavy model retraining.

\noindent
\textbf{Non-agent Approaches.}
Non-agent video reasoning methods aim to improve reasoning capability without introducing external tools or auxiliary models.
An early line of work focuses on chain-of-thought supervised fine-tuning, which teaches models to generate explicit intermediate reasoning traces directly from annotated examples~\cite{zhang2025video-cot,ghazanfari2025chain}.
More recent studies extend this line through post-training, most commonly reinforcement learning and preference optimization, to encourage more faithful, structured, and grounded reasoning traces~\cite{feng2025video-r1,wang2025videorft,li2025veripo,park2025deepvideo,huang2025vistadpo,dang2025twgrpo}.
Overall, rather than relying on iterative planning and multi-step agentic execution, non-agent methods improve reasoning mainly by moving from CoT supervision to stronger post-training objectives.

Following the R1-style rule-based reinforcement learning paradigm, a large body of recent work builds upon GRPO and adapts it to video-specific reasoning challenges.
Video-R1~\cite{feng2025video-r1} is one of the representative starting points of this line and introduces temporal-aware GRPO by contrasting rewards between temporally ordered and shuffled frames, encouraging the model to rely on correct temporal relationships.
TW-GRPO~\cite{dang2025twgrpo} further refines the optimization by using token-weighted GRPO, emphasizing informative reasoning tokens and employing soft, multi-level rewards.
It can reduce training variance and overly long or redundant reasoning traces.
DeepVideo-R1~\cite{park2025deepvideo} reformulates GRPO as advantage regression, removing clipping-style constraints and pairing it with difficulty-aware data augmentation to maintain informative rewards.
Keye-VL 1.5~\cite{yang2025kwai} adopts GSPO for verifiable reward-based post-training to better handle sparse and high-variance rewards in video reasoning, and further introduces progressive hint sampling to improve rollout efficiency on hard samples.

An alternative line of work replaces or complements reinforcement learning rollouts with preference-based optimization, typically offering improved stability and sample efficiency.
VistaDPO~\cite{huang2025vistadpo} proposes hierarchical preference optimization at instance, temporal, and perceptive levels, aligning language responses with events and objects through fine-grained supervision such as timestamps and bounding boxes.
video-SALMONN-o1~\cite{sun2025video-salmonn-o1} introduces process-level preference optimization that enables step-aware alignment without relying on an explicit reward model.
VerIPO~\cite{li2025veripo} further bridges GRPO-based learning and preference optimization through an iterative loop, where verifier-filtered rollouts are converted into high-quality contrastive data for subsequent preference-based training.

Beyond modifying the optimization algorithm, many works focus on designing video-specific reward signals for the features of video understanding tasks.
VideoRFT~\cite{wang2025videorft} introduces a semantic consistency reward that aligns the textual reasoning trace with visual features, directly penalizing visually ungrounded narratives.
Time-R1~\cite{wang2025time} targets temporal grounding and designs verifiable rewards based on temporal IoU with deviation-aware penalties, replacing rigid supervised penalties in SFT.
VidBridge-R1~\cite{chen2025vidbridger1bridgingqacaptioning} bridges video QA and captioning by designing proxy tasks, including DarkEventInfer and MixVidQA, and uses them to construct task-aligned rewards.
VideoCap-R1~\cite{meng2025videocap} proposes caption-specific reward modeling.
It evaluates whether key subjects, actions, and attributes are correctly identified during structured reasoning, and measures factual event coverage in the generated caption.

In addition to reinforcement learning and preference optimization, several works improve textual video reasoning by directly fine-tuning models on high-quality chain-of-thought annotations.
Chain-of-Frames~\cite{ghazanfari2025chain} trains models with frame-aware reasoning traces that explicitly reference relevant frames, enabling grounded reasoning without auxiliary frame selectors or other tools.
Video-CoT~\cite{zhang2025video-cot} provides large-scale chain-of-thought data that enables supervised fine-tuning to enhance temporal and spatial reasoning.

Spatial intelligence has also become an increasingly important component of video reasoning. It remains challenging due to dynamic object layouts, viewpoint changes, and the need to maintain spatial consistency over time.
SpaceR~\cite{ouyang2025spacer} represents object locations and relations in an explicit spatial map and trains models with GRPO to reason over such map-based spatial representations.
vsGRPO~\cite{liao2025improved} adopts an R1-Zero-like reinforcement learning scheme to directly optimize visual-spatial reasoning behaviors.
VIDEO-STR~\cite{wang2025video-str} models spatial and temporal interactions using object-centric relation graphs, enabling structured reasoning over multi-object layouts across time.
SpatialLadder~\cite{li2025spatialladder} proposes a progressive curriculum that incrementally builds spatial reasoning from basic perceptual grounding to higher-level spatial abstraction.
Cambrian-S~\cite{yang2025cambrian} targets long-horizon spatial cognition by introducing visual spatial recall and continual visual spatial counting tasks, emphasizing sustained spatial memory over extended videos.
Overall, these models enhance spatial modeling, enabling video MLLMs to better capture real-world spatial structure.

\subsubsection{Thinking with Videos}
\label{sec:reason_o3_like}

In text-only reasoning, models may ignore important visual cues and produce hallucinated statements~\cite{feng2025video-r1,wang2025videorft}.
These statements can stray from the video content and lead to incorrect answers.
Readers also cannot easily verify whether long chains of thought are grounded or unfounded, which limits interpretability.
For these reasons, rechecking the video and strengthening spatio-temporal grounding during reasoning is important.
Many recent works in 2025~\cite{zhang2025thinking,yan2025videochat,zhang2025rewatch,wang2025pixel,Wen2025-busterx, ouyang2025conan,meng2025open,Li2026-OmniFake, wang2025video-thinker,fu2025love,yang2025longvt,gu2025thinking} draw inspiration from OpenAI-o3's ``thinking with images'' and extend this idea to videos to implement this paradigm.
Specifically, thinking with videos refers to a reasoning paradigm in which the model interleaves textual reasoning with explicit visual grounding.
The model, much like a human, can decide when to look back at the video and where to focus, and incorporate retrieved evidence into its reasoning.
This mechanism improves the readability and reliability of the reasoning trace, reduces hallucination, and has been shown to yield better performance than pure text-only reasoning.
Technically, most methods implement this paradigm through structured tool usage or structured output formats.
These approaches can still be grouped into two categories: \textbf{agentic} and \textbf{non-agent}.

\noindent
\textbf{Agentic Approaches.}
Agent-based approaches~\cite{zhang2025thinking,yan2025videochat,zhang2025rewatch,wang2025pixel,ouyang2025conan,he2025framethinker,yuan2025videoexplorerthinkvideosagentic,wu2025reinforcing,ge2025framemind,zeng2026videoo3,liu2026videotemp,meng2025cyberv,wang2025active,lin2026videoseek} treat the model as a controller that dynamically manages the flow of reasoning. At each step, the agent decides whether to provide a direct answer or to invoke a perception tool to gather additional visual evidence. 
Tools can include retrieving temporal segments or specific frames, cropping spatial regions, or drawing spatial annotations.
The returned visual tokens are incorporated into the model's internal state for iterative reasoning.

Some works use pure reinforcement learning to train the agent's tool-usage policy for multi-round reasoning that interleaves text and visual evidence gathering. Reinforcement rewards are typically designed to reflect the quality of tool utilization and the temporal or spatial grounding efficacy during reasoning. 
For example, VideoChat-R1.5~\cite{yan2025videochat} learns to localize both temporal intervals and spatial bounding boxes with RL where the clue reward measures alignment between predictions and ground truth, enabling effective temporal and spatial localization during question answering. 
FrameMind~\cite{ge2025framemind} provides dual tools for coarse temporal scanning and fine spatial inspection and proposes DRFS-FRPO to encourage the low-resolution path to trigger tool calls when appropriate. 
These methods demonstrate that agentic RL yields adaptive evidence acquisition strategies tailored to task demands.

Beyond pure RL, many recent methods combine supervised fine-tuning with reinforcement learning to leverage high-quality chain-of-thought (CoT) data for cold-start training and use RL to further refine evidence acquisition policies. 
VITAL~\cite{zhang2025thinking} constructs 72k CoT datasets using Gemini to annotate step-by-step temporal reasoning and tool invocation over long videos. And it applies difficulty-aware group-relative policy optimization to mitigate task imbalance in RL.
Pixel Reasoner~\cite{wang2025pixel} introduces pixel-space reasoning traces to supervise atomic visual actions. And a curiosity-driven RL objective then encourages tools like select-frame and zoom-in in pixel space.
ViLaSR~\cite{wu2025reinforcing} adopts a spatial drawing paradigm that highlights structural cues via bounding boxes and auxiliary lines, also using reflective rejection sampling to enhance correction reasoning in the SFT stage.
Overall, the SFT-CoT cold start scheme provides structured reasoning priors and interpretable tool usage. RL then optimizes adaptive decision-making and the integration of visual evidence.

Moreover, several works emphasize training-free paradigms.
CyberV~\cite{meng2025cyberv} observes that long chains of thought may cause visual attention to drift and introduces a cybernetic system. It adaptively inserts key frames during inference, reducing reliance on annotated reasoning traces.
AVP~\cite{wang2025active} models long-video understanding as an active plan-observe-reflect process. The agent also acquires visual evidence selectively and reduces computation compared to caption-based agentic methods.
More recently, VideoSeek~\cite{lin2026videoseek} introduces a think-act-observe loop with a multi-granular toolkit to actively seek answer-critical evidence, achieving strong long-horizon reasoning performance with substantially fewer viewed frames.

\noindent
\textbf{Non-agent Approaches.} Non-agent methods~\cite{meng2025open,wang2025video-thinker,zhang2025rewatch,maaz2025videor2} aim to directly generate a grounded reasoning trace that is natively verifiable, without invoking any external tool functions. These models produce step-by-step textual reasoning while explicitly exposing spatio-temporal evidence (e.g., timestamps, object references, bounding boxes, or other observations) within the reasoning process itself.
This is typically enforced through a structured output schema, and the model is trained (often via a CoT cold start followed by reinforcement learning) to jointly satisfy answer correctness and evidence-format compliance.

Open-o3-Video~\cite{meng2025open} formulates grounded video reasoning as structured generation of explicit spatio-temporal evidence. The model can produce answers together with concrete timestamps and spatial boxes in the reasoning process.
Video-Thinker~\cite{wang2025video-thinker} learns thinking with videos without tool calls by directly producing cues such as explicit temporal markers and query-conditioned captions that guide subsequent reasoning.

Overall, the non-agent paradigm of thinking with videos emphasizes avoiding complex tool invocation and multi-round interaction, instead enabling the model to natively retrieve and present evidence from the video within a single reasoning process.
By directly coupling reasoning with explicit, video-grounded outputs, this approach offers a lighter, more efficient alternative. And this direction still leaves substantial room for further exploration and improvement.

\section{Subfields: Various Video Types}
\label{sec:subfields}

In this section, we also review several specific video types and applications, including egocentric videos, sports videos, instructional videos, medical videos, and movies. We review each category in detail.

\subsection{Egocentric Videos}

Egocentric video understanding~\cite{grauman2022ego4d} shifts the research focus from passive third-person observation to active, first-person embodied engagement, requiring models to interpret interactions, intentions, and 4D spatio-temporal dynamics from the wearer's perspective. 
Before the recent rise of MLLMs, this area had already developed important foundations in egocentric anticipation, video modeling, and video-language pretraining, such as Anticipative Video Transformer~\cite{girdhar2021anticipative}, TimeSformer~\cite{bertasius2021space}, EgoVLP~\cite{lin2022egocentric}, and EgoVLPv2~\cite{pramanick2023egovlpv2}.

Recent advancements have significantly refined the granularity of perception and the depth of reasoning in this domain, where they build the new data engines and benchmarks.
Earlier works focus on enhancing fine-grained grounding and event representation. 
For example, EgoMask~\cite{liang2025fine} establishes a pixel-level benchmark for precise spatio-temporal grounding, while DMC3~\cite{zou2025dmc3} employs dual-modal counterfactual contrastive learning to mitigate hallucinations in interaction understanding. 
Moving beyond static perception to complex spatio-temporal reasoning, recent works explore RL-based methods in this direction.
ST-Think~\cite{wu2025st} and VLN-R1~\cite{qi2025vln} leverage Reinforcement Learning (RL) to master 4D world modeling and vision-language navigation, respectively. 
To tackle the challenge of ultra-long contexts, Ego-R1~\cite{tian2025ego} introduces a ``Chain-of-Tool-Thought'' mechanism that dynamically coordinates hierarchical retrieval and tool usage for week-long video reasoning.

The field is further evolving towards proactive and socially aware systems. VideoLLM-EyeWO~\cite{zhang2025eyes} proposes a proactive video-LLM capable of determining \textit{when} to speak in streaming scenarios, a capability extended by EgoSocial~\cite{wang2025egosocialbenchmarkingproactiveintervention} to social intervention timing in AR/VR environments. 
In safety-critical domains, DVBench~\cite{zeng2025visionllmsroadreadycomprehensive} evaluates the robustness of these models in driving scenarios, emphasizing the need for reliable spatio-temporal causal reasoning.

\subsection{Sports Videos}

Sports video understanding involves fast, fine-grained actions, frequent camera cuts (e.g., multi-camera switching and replays), and domain-specific rules and terminology.
As a result, key evidence is often highly time-localized, and correct reasoning requires both accurate temporal grounding and sports knowledge.
Recent MLLM-based approaches~\cite{xia2024sportu,rao2025towards,jiang2025domain,zou2025deepsport,chen2025finequest,bao2025tennistv,rai2026learning} mainly focus on two directions: building domain-aligned datasets for rule- and tactic-aware reasoning, and improving evidence acquisition through temporal localization or structured intermediate representations.

For the first direction, SPORTU~\cite{xia2024sportu} formalizes multi-level sports reasoning evaluation, emphasizing the gap between general MLLM perception and rule-oriented decision making.
Unisoccer~\cite{rao2025towards} scales up soccer-centric multimodal data and trains a unified soccer foundation encoder to support heterogeneous downstream tasks.
Jiang et al.~\cite{jiang2025domain} provide a practical curriculum-style recipe with short event clips to robustly adapt a general video VLM to soccer-specific QA and classification.
For the second direction, DeepSport~\cite{zou2025deepsport} proposes an agentic think-with-videos loop that refines temporal evidence retrieval, targeting the sports-specific failure mode where sparse uniform sampling misses brief but decisive events.
And FineQuest~\cite{chen2025finequest} enhances training-free sports VideoQA by grounding visual evidence into a sports knowledge scene graph, enabling dual-mode structured reasoning that is robust to rapid actions and frequent camera cuts.

Despite recent progress, models still struggle to precisely localize decisive moments and consistently apply sports rules.
Future work may focus on generating explicit spatio-temporal evidence aligned with rules, as well as improving generalization across leagues and broadcast styles.

\subsection{Instructional Videos}

Instructional (course) videos, such as lectures and tutorials, are typically long and information-dense, with tight coupling between speech and visually grounded content, including slides, equations, and step-wise demonstrations.
Unlike open-domain videos, their core challenges lie less in action recognition and more in tracking procedural progress, aligning document-centric visual evidence with narration over time, and evaluating whether models genuinely acquire and transfer knowledge.
Recent MLLM-based studies~\cite{hu2025video,song2025video,zhao2025noteit} share several common technical directions.

A first line of work focuses on evaluation protocols that explicitly measure learning-oriented abilities in instructional videos.
For example, Video-MMMU~\cite{hu2025video} and Video-MMLU~\cite{song2025video} introduce lecture-focused benchmarks that treat video understanding as knowledge acquisition under perception and reasoning constraints.
Meanwhile, InstructionBench~\cite{wei2025instructionbench} benchmarks temporally ordered and procedurally structured instructional reasoning.
Another line of work emphasizes evidence selection and cross-modal integration under limited perception budgets.
For example, DocVideoQA~\cite{wang2025docvideoqa} focuses on document-centric instructional videos and studies temporal alignment and fusion between dense on-screen text and narration.
More recent efforts move beyond short-form question answering toward structured knowledge extraction and instructional assistance.
For example, NoteIt~\cite{zhao2025noteit} converts instructional videos into hierarchical and interactable notes, enabling reusable knowledge representations.
InsTALL~\cite{nguyen2025install} models instructional procedures as task graphs to support progress tracking and next-step prediction.

Overall, understanding instructional videos enables cross-modal knowledge alignment and process-aware modeling for educational applications.
Future work may focus on personalized learning agents built on instructional video corpora or multilingual course agent development.

\subsection{Medical Videos}

Medical video understanding is a high-stakes subfield of long video understanding, with long procedures, subtle visual changes, and strong procedural and domain constraints.
Compared with generic videos, it requires both global procedural context and local anatomical or tool-related evidence, together with stable temporal modeling.

Before multimodal large models, most studies focus on task-specific surgical video analysis, such as phase and step recognition, fine-grained ``instrument--verb--target interaction'' modeling, and skill assessment~\cite{czempiel2020tecno,ramesh2021multi,nwoye2022rendezvous,nwoye2023cholectriplet2022}.
These works provide strong medical priors, but they mainly remain within specialized recognition pipelines.

Recent work increasingly introduces vision-language pretraining into this domain.
Surgery-specific self-supervision improves transfer across downstream tasks~\cite{ramesh2023dissecting}.
SurgVLP~\cite{yuan2025learning} learns from narrated surgical video lectures, SurgVISTA~\cite{yang2026large} extends this line to video-level self-supervised pretraining with joint spatio-temporal modeling, and MM-OR~\cite{ozsoy2025mm} broadens the setting to multimodal operating-room streams with RGB-D video, audio, speech transcripts, and robot logs.

Medical VLMs and MLLMs further push the field from recognition to reasoning.
LLaVA-Surg~\cite{li2024llavasurg}, Surgical-LLaVA~\cite{jin2024surgical-llava}, EndoChat~\cite{wang2025endochat}, SurgVLM~\cite{zeng2025surgvlm}, SurgVidLM~\cite{wang2025surgvidlm}, and SurgViVQA~\cite{drago2025surgvivqa} adapt large vision-language models to surgical dialogue, multi-task understanding, multi-grained video reasoning, and temporally grounded VideoQA.
These studies show a clear trend toward richer language supervision, longer temporal context, and more explicit reasoning over medical video evidence.

Beyond surgery, recent multimodal models also extend to non-surgical continuous imaging streams such as ultrasound.
EchoCLIP~\cite{christensen2024vision} learns vision-language representations for echocardiogram interpretation.
MMSummary~\cite{guo2024mmsummary} explores multimodal summary generation for fetal ultrasound video.
Sonomate~\cite{guo2026visually} further builds a visually grounded language model for fetal ultrasound understanding with video-text alignment and VQA.

Overall, the field is moving from specialized surgical recognition to medical VLMs and MLLMs with stronger multimodal pretraining and more explicit evidence-based reasoning.
However, current models still struggle with rare events, cross-domain transfer, and clinically faithful explanation.
Future progress may depend on better integration of medical knowledge, temporal memory, and multimodal evidence.




\subsection{Movie and Narrative Videos}

Movie and narrative videos pose a distinct challenge for long video understanding.
Their key evidence is often scattered across scenes, and correct answers depend on plot progression, character dynamics, and causal links rather than local visual cues.
Early movie-oriented resources, including MovieQA~\cite{tapaswi2016movieqa}, MovieNet~\cite{huang2020movienet}, MAD~\cite{soldan2022mad}, MoVQA~\cite{zhang2023movqa}, and MovieChat~\cite{song2024moviechat}, lay the foundation for long-form movie understanding.

Recent work increasingly focuses on narrative-centric evaluation.
SFD~\cite{ghermi2024short} reduces shortcut and data-leakage issues with longer, public movie-style videos.
SCVBench~\cite{you2025scvbench} and VRBench~\cite{yu2025vrbench} evaluate story understanding through event ordering, multi-turn decomposition, and temporally grounded multi-step reasoning.
SeriesBench~\cite{zhang2025seriesbench}, Cin\'easte~\cite{shah2025cin}, and MovieCORE~\cite{faure2025moviecore} extend this line to series-level plot tracking, fine-grained contextual movie QA, and deeper cognitive reasoning.

Beyond evaluation, recent studies also explore more explicit narrative structure.
SCVBench~\cite{you2025scvbench} introduces StoryCoT to decompose story understanding into event-level reasoning steps, while SeriesBench~\cite{zhang2025seriesbench} uses PC-DCoT to organize evidence along both plot and character chains.
For longer videos, ARC-Chapter~\cite{pu2025arc} organizes hour-long content into navigable chapters and hierarchical summaries, making long-range narrative structure more accessible.
These designs make long-range narrative evidence more accessible for reasoning and reduce reliance on local visual shortcuts.

Overall, the field moves from clip-level perception to explicit narrative modeling and structured story reasoning.
However, current models still struggle with temporally dispersed evidence, cross-scene character tracking, and consistent causal explanation.
Future progress may depend on stronger narrative memory, better audio-dialogue grounding, and more explicit retrieval of long-range story evidence.




\section{Datasets and Benchmarks}
\label{sec:benchmark_results}

\subsection{Common Training Datasets}
\label{sec:train_data}

We first present an overview of the large-scale datasets used to train video MLLMs, including those for instruction tuning and reinforcement learning.
To better reflect the diversity of supervision in current video understanding, we categorize these datasets by task type.
Accordingly, we summarize representative datasets for \textbf{Video QA}, \textbf{Video Captioning}, \textbf{Video Temporal Grounding}, and \textbf{Long Video Memory}, with a focus on their video duration, covered modalities, and annotation formats.

\begin{table*}[t]
\centering
\renewcommand{\arraystretch}{1.2}
\caption{Representative training datasets for video MLLMs (Sec.~\ref{sec:train_data}). ``Scale'' refers to the number of (video clip, text) pairs by default, and marked entries report the number of videos instead.}
\scalebox{0.8}{
\begin{tabular}{p{3.6cm} p{0.5cm} p{13cm} p{3cm}}
\toprule
\textbf{Dataset} & \textbf{Year} & \textbf{Focus} & \textbf{Scale} \\
\midrule

\multicolumn{4}{l}{\textit{\textbf{I. Video QA}}} \\
\midrule
VideoChat2-IT~\cite{MVBench} & 2024  & Mixed video instruction tuning across diverse video tasks. & 1.9M \\
LLaVA-Video-178K~\cite{zhang2024llavavideo} & 2024  & Detailed captioning, open-ended QA, and multiple-choice QA for video instruction tuning. & 1.3M \\
VideoCoT~\cite{wang2024videocot} & 2024  & Video QA with explicit reasoning rationales for open-ended and multiple-choice questions. & 22K \\
VideoEspresso~\cite{han2025videoespresso} & 2025  & Video QA with multimodal intermediate evidence and core-frame selection. & 202K \\
Video-R1~\cite{feng2025video-r1} & 2025  & Reinforced video reasoning with CoT supervision and RL post-training. & 165K CoT + 260K RL \\
VideoRFT~\cite{wang2025videorft} & 2025  & Reinforced fine-tuning for video reasoning. & 102K CoT + 310K RL  \\
LongVideo-Reason~\cite{chen2025scaling} & 2025 & Long-video reasoning with SFT and RL training splits. & 52K \\
STGR~\cite{meng2025open} & 2025  & Video QA with explicit timestamps and bounding boxes in reasoning traces. & 30K CoT + 36K RL \\
ReWatch-CoT~\cite{zhang2025rewatch} & 2025  & Multi-agent ReAct-style data with repeated observation and retrieval steps for re-watching. & 135K \\
VideoZoomer~\cite{ding2025videozoomer} & 2025  & Multi-round temporal zooming trajectories for tool-based evidence focusing. & 11K \\
VideoSIAH~\cite{yang2025longvt} & 2025  & Tool-integrated SFT and RL data for clip cropping, rethinking, and long-video QA. & 247.9K \\
Conan~\cite{ouyang2025conan} & 2025  & Agent-style multi-scale evidence search with frame identification and action decision. & 91K \\
Seeker-173K~\cite{zeng2026videoo3} & 2026  & Multi-turn tool-interaction data for clue seeking, fine inspection, and adaptive stopping. & 173K \\
LongVideo-R1~\cite{qiu2026longvideo-r1} & 2026  & Multi-round CoTwT navigation traces for long-video reasoning with tool use. & 33K \\

\midrule
\multicolumn{4}{l}{\textit{\textbf{II. Video Captioning}}} \\
\midrule
Panda-70M~\cite{chen2024panda} & 2024  & Large-scale video-text supervision through automatically selected captions. & 70M \\
ShareGPT4Video~\cite{chen2024sharegpt4video} & 2024  & High-quality dense captions for video understanding and generation. & 4.8M\\
Video ReCap~\cite{islam2024video} & 2024  & Recursive multi-level captioning from clip-level to global summaries. &  5.3M\\
Vript~\cite{yang2024vript} & 2024  & Structured dense captions with scene and narrative progression. &  420K \\
MiraData~\cite{ju2024miradata} & 2024  & Long-video captions with scene splits, camera language, and metadata. & 330K \\
FineVideo~\cite{farre2024finevideo} & 2024  & Structured video captions for video-language and generative modeling. &  43.8K (Videos) \\
Tarsier2-Recap-585K~\cite{yuan2025tarsier2} & 2025  & High-quality recaptioning for fine-grained video description and alignment. & 585K \\
UltraVideo~\cite{xue2025ultravideo} & 2025  & High-quality UHD video captioning supervision. & 58.8K \\
HMD-270K~\cite{zhong2025owlcap} & 2025  & Motion-detail balanced captions for human-centric video understanding. & 270K \\
TimeChatCap-42K~\cite{yao2026timechat} & 2026  & Time-aware and structural audio-visual video scripting. & 42K \\

\midrule
\multicolumn{4}{l}{\textit{\textbf{III. Video Temporal Grounding}}} \\
\midrule
TimeIT~\cite{ren2023timechat} & 2023  & Unified instruction tuning for temporal grounding tasks. & 125K \\
VTimeLLM Data~\cite{huang2024vtimellm} & 2024  & Three-stage temporal instruction curriculum for boundary-aware video LLMs. & 134K (Videos) \\
VTG-IT-120K~\cite{guo2025vtgllm} & 2025  & High-quality temporal grounding instruction data with standardized time tokens. & 120K \\
E.T. Instruct 164K~\cite{liu2024etbench} & 2024  & Event-level instruction data for fine-grained temporal understanding. & 164K \\
TimePro~\cite{zeng2025timesuite} & 2025  & Temporal grounding and temporal grounded captioning. & 349K \\
Moment-10M~\cite{qian2024momentor} & 2024  & Large-scale moment-level instruction tuning for single- and cross-segment tasks. & 10.4M \\
Vid-Morp~\cite{bao2024vidmorp} & 2024  & Query-boundary pseudo-labels without human cleaning. & 200.3K \\
VideoITG~\cite{wang2025videoitg} & 2025  & Scalable temporal grounding annotation for video MLLMs. & 500K \\
TimeLens-100K~\cite{zhang2025timelens} & 2025  & High-precision temporal grounding data with multi-step validation. & 100K \\
MTVR~\cite{zhang2025thinking} & 2025  & Multi-turn temporal grounding and QA with clip-cropping tool calls. & 72K SFT + 110K RL \\

\midrule
\multicolumn{4}{l}{\textit{\textbf{IV. Long Video Memory}}} \\
\midrule
VideoMarathon~\cite{lin2025unleashing} & 2025  & Hour-scale instruction data for long-video memory and long video-language understanding. & 3.3M \\
M3-Agent~\cite{long2026seeing} & 2026  & Entity-centric multimodal long-term memory for agentic video understanding. & 10.9K (Videos) \\
\bottomrule
\end{tabular}
}
\label{tab:train_data}
\end{table*}

\noindent
\textbf{Video QA.}
Pre-MLLM VideoQA mostly relies on widely used datasets such as MSVD-QA~\cite{xu2017msvdqa}, TGIF-QA~\cite{jang2017tgif}, ActivityNet-QA~\cite{yu2019activitynetqa}, TVQA~\cite{lei2018tvqa}, NExT-QA~\cite{xiao2021nextqa}, and CLEVRER~\cite{yi2019clevrer}.
These datasets typically provide both training and test splits, and most of them supervise short answers over relatively short videos.
HowToVQA69M~\cite{yang2021howtovqa} serves as an early bridge to scale, with 69M automatically generated video-question-answer triplets from narrated videos.

With video MLLMs, VideoInstruct100K~\cite{maaz2024videoinstruct100k} introduces 100K video-instruction pairs for conversational tuning, using human-assisted and semi-automatic annotation to cover description, summarization, QA, and dialogue.
VideoChat2-IT~\cite{MVBench} scales instruction tuning to 1.9M samples from 34 sources, and it unifies many earlier video tasks into one mixed instruction corpus.
LLaVA-Video-178K~\cite{zhang2024llavavideo} contributes 178,510 videos and about 1.3M instruction samples, with synthetic detailed captions, open-ended QA, and multiple-choice QA over dynamic untrimmed videos.
These datasets shift the focus from task-specific supervision to broad video instruction following.

Subsequently, many CoT training datasets emerge to further enhance video reasoning ability.
VideoCoT~\cite{wang2024videocot} contains 11K videos and 22K QA items, and it provides active-annotation CoT rationales for both open-ended and multiple-choice QA.
VideoEspresso~\cite{han2025videoespresso} is a large automatic VideoQA dataset with multimodal intermediate evidence and core-frame selection, and it is useful for training models to reason over selected visual evidence.
Video-R1-CoT-165K~\cite{feng2025video-r1} and VideoRFT-CoT-102K~\cite{wang2025videorft} are large-scale cold-start CoT datasets for video reasoning. They are paired with Video-R1-260K and VideoRFT-RL-310K, respectively, to support subsequent reinforcement learning.
LongVideo-Reason~\cite{chen2025scaling} provides about 52K long-video question-reasoning-answer pairs, with about 18K samples used for SFT and the rest supporting RL.
STGR-CoT-30K~\cite{meng2025open} adds explicit timestamps and bounding boxes to each reasoning trace, making it useful for grounded spatio-temporal reasoning.

In more recent thinking-with-videos methods, a series of datasets are introduced for multi-round reasoning.
MTVR-CoT-72K~\cite{zhang2025thinking} from VITAL is a tool-augmented multi-task dataset for QA and temporal grounding, and it explicitly supports on-demand visual sampling during reasoning.
ReWatch-CoT-135K~\cite{zhang2025rewatch} uses a Multi-Agent ReAct pipeline over detailed captions, so its traces contain repeated observation and retrieval steps that simulate re-watching.
VideoZoomer~\cite{ding2025videozoomer} uses about 11K exemplar and reflection trajectories to teach a \texttt{<video\_zoom>} tool, making it a clear multi-round tool-use dataset.
VideoSIAH~\cite{yang2025longvt} provides 247.9K tool-integrated SFT samples, plus RL and RFT data, and it trains native clip-cropping and rethinking loops for long videos.
Conan-91K~\cite{ouyang2025conan} records frame identification, evidence reasoning, and action decision, and it supports agent-style reasoning over multi-scale visual evidence.
Seeker-173K~\cite{zeng2026videoo3} from Video-o3 is a native multi-turn tool-interaction corpus built for clue seeking, fine inspection, and adaptive stopping.
LongVideo-R1~\cite{qiu2026longvideo-r1} adds 5.6K CoTwT trajectories over long videos; these are multi-round navigation traces with an average of 5.8 steps, and they are later expanded into about 33K SFT samples.

Overall, Video QA data shifts from short-answer supervision, to large-scale instruction tuning, and then to reasoning data, including one-shot CoT and agentic multi-round trajectories.

\noindent
\textbf{Video Captioning.}
Video captioning data has gone through three fairly distinct phases: early clip-level and dense-captioning benchmarks, large-scale auto-captioned and recaptioned corpora, and more recent resources that push captions toward grounded, audio-visual, and time-aware scripts.

The early phase is defined by a handful of benchmarks that fix the task's basic forms. MSR-VTT~\cite{xu2016msr} and VATEX~\cite{wang2019vatex} target open-domain clip-level captioning, while ActivityNet Captions~\cite{krishna2017dense}, YouCook2~\cite{zhou2018towards}, TVC~\cite{lei2020tvr}, ViTT~\cite{huang2020multimodal}, and Ego4D narrations~\cite{grauman2022ego4d} extend annotation to temporally localized, procedural, subtitle-aware, timeline-tagged, or egocentric descriptions. TVC is the captioning counterpart of the TVR video-subtitle retrieval resource, and ViTT and Ego4D narrations sit closer to dense language supervision than to conventional caption-only benchmarks. Captions here are mostly single-sentence, event-level, or narration-level summaries, but the two settings that later work keeps inheriting---short clip captioning and dense long-video narration---are already in place.

Since 2024, the dominant effort is to scale caption supervision through automatic relabeling, recaptioning, and richer descriptions. Panda-70M~\cite{chen2024panda} pushes video-text supervision to 70M clips with captions selected by cross-modality teachers, taking a scale-first route. ShareGPT4Video~\cite{chen2024sharegpt4video} and, for hour-long videos, Video ReCap~\cite{islam2024video} with its Ego4D-HCap multi-level summaries extend this line to longer and more hierarchical descriptions. A parallel, structure-first thread shows up in Vript~\cite{yang2024vript}, MiraData~\cite{ju2024miradata}, and FineVideo~\cite{farre2024finevideo}, which enrich captions with scene splits, narrative progressions, camera language, speech-aligned metadata, or other structured long-video annotations. Tarsier2-Recap-585K~\cite{yuan2025tarsier2}, UltraVideo~\cite{xue2025ultravideo}, and HMD-270K~\cite{zhong2025owlcap} continue this trend with higher-quality or more specialized recaptioned corpora; Tarsier itself is better viewed as a video description model and training/evaluation recipe~\cite{wang2024tarsier}, with Tarsier2-Recap-585K as its data-side contribution. At this stage, caption data is no longer only a benchmark target but a general supervision source for video-language and text-to-video models.

More recent work turns captioning into something closer to structured video scripting, along two largely complementary directions. On the grounding side, ViCaS~\cite{athar2025vicas} and HowToGround1M/iGround~\cite{kazakos2025large} tie object mentions in captions to dense boxes or masks, and PerceptionLM~\cite{cho2025perceptionlm} pushes this further on the training-data side by releasing PLM-Video-Auto and PLM-Video-Human, which together cover synthetic video captions/QA as well as human-annotated region-level, dense, and spatio-temporally grounded video captioning supervision. On the omnimodal and time-aware side, UGC-VideoCap~\cite{wu2025ugc} treats audio as part of caption semantics rather than auxiliary context, and OmniDCBench together with TimeChatCap-42K~\cite{yao2026timechat} reformulates captioning as omni dense captioning with explicit timestamps and multi-dimensional audio-visual scripts.

\noindent
\textbf{Video Temporal Grounding.}
Video temporal grounding (VTG) data has evolved from web-scale pretraining, to unified instruction tuning, and most recently to reasoning- and tool-oriented corpora.

Early work emphasizes scale. YT-Temporal-180M~\cite{zellers2021merlot} collects 6M YouTube videos and 180M short clips, pairing each with ASR transcripts aligned via dynamic time warping, establishing a fully automatic pretraining template reused by many later datasets.

With the rise of video MLLMs, several works convert heterogeneous grounding benchmarks into unified instruction formats. TimeIT~\cite{ren2023timechat} aggregates 12 benchmarks into 125K samples over six tasks (e.g., dense captioning, moment retrieval, highlight detection). VTimeLLM~\cite{huang2024vtimellm} adopts a three-stage curriculum covering feature alignment, boundary awareness, and dialogue tuning. VTG-IT-120K~\cite{guo2025vtgllm} emphasizes annotation quality, re-labeling 51.9K low-quality TimeIT samples with Gemini-1.5 Pro and standardizing absolute time tokens. E.T. Instruct 164K~\cite{liu2024etbench} broadens coverage to nine event-level tasks from 14 sources. TimePro~\cite{zeng2025timesuite} scales grounded tuning to 349K annotations and newly introduces temporal grounded captioning. At the upper end, Moment-10M~\cite{qian2024momentor} uses an automated instance--event engine to produce 10.4M clip-level instructions over 64.9K long videos (avg. 403s), covering single- and cross-segment tasks.

A parallel line targets data diversity and precision. Vid-Morp~\cite{bao2024vidmorp} mines 52.7K in-the-wild videos and uses GPT-4o to generate 200.3K query--boundary pseudo-labels without human cleaning. VideoITG~\cite{wang2025videoitg} introduces the VidThinker pipeline (chunk retrieval plus frame-level classification) to produce 500K grounding annotations over 40K videos. TimeLens-100K~\cite{zhang2025timelens} re-annotates 20K videos with Gemini-2.5 Pro through a four-step validation procedure, uniformly covering 0--240s durations.

Most recently, VTG data has shifted toward reasoning supervision. ActivityNet-RTL~\cite{huang2024lita} uses GPT-4 to build 33.5K reasoning-style ``when'' questions with rationale answers. Following R1-style post-training, TimeRFT~\cite{wang2025time} filters 339K samples into 2.5K medium-difficulty instances via IoU-based difficulty modeling, and TVG-R1~\cite{chen2025datasets} splits seven VTG datasets by IoU into a 13K cold-start SFT set and an 18K RL set with temporal chain-of-thought. VTTS-80K~\cite{yan2025videochat} unifies QA, temporal grounding, and spatial grounding into a single thinking-trace format. Going further, MTVR~\cite{zhang2025thinking} introduces tool-augmented CoT data (72K for SFT, 110K for RL), where Gemini-2.5 Pro annotates multi-turn \texttt{<tool\_call>} traces that invoke clip-cropping tools during reasoning.

Overall, VTG datasets have progressed from web-scale pretraining, to multi-task instruction tuning, to precision-oriented re-annotation, and finally to reasoning- and tool-oriented corpora that supervise chain-of-thought, verifiable rewards, and multi-turn tool use.

\noindent
\textbf{Long Video Memory.}
For memory-augmented long video understanding methods, training data primarily serve two purposes: fine-tuning the visual-linguistic alignment module (typically a Q-Former or cross-attention mechanism) and training agentic models to perform tool-invoked memory retrieval and reasoning. Most of the existing works rely on publicly available datasets with only minor cleaning and filtering. Only two recent efforts have proposed dedicated large-scale training sets tailored for long-video memory modeling.

The first is VideoMarathon~\cite{lin2025unleashing}, designed to address the scarcity of hour-scale video instruction data. It comprises approximately 9,700 hours of video (28K videos, 3--60 minutes each) and 3.3 million QA pairs across six dimensions (temporality, spatiality, object, action, scene, and event). Videos are sourced from five public datasets and filtered to retain only those with at least three distinct events. A hierarchical captioning pipeline (clip-level via Qwen2VL-7B, event- and global-level via DeepSeek-V3) produces multi-granularity descriptions, from which topic-specific prompts generate diverse QA pairs spanning 22 tasks in both open-ended and multiple-choice formats.

The second is the companion training set of M3-Agent~\cite{long2026seeing}, aimed at constructing entity-centric multimodal long-term memory for agentic video understanding. It contains 500 long videos (26,943 thirty-second clips) with 10,952 synthesized memory demonstrations and 2,736 QA pairs. Memory annotations are produced through a three-stage hybrid pipeline: episodic memories are synthesized by jointly prompting GPT-4o and Gemini-1.5-Pro, cross-modal identity equivalences are established via an automated meta-clip mining algorithm that pairs faces with voices, and semantic memories (character attributes, relationships, contextual knowledge) are extracted through a similar hybrid strategy. The memorization model is trained via SFT on these demonstrations, while the control policy is further optimized with DAPO reinforcement learning using binary correctness rewards.

\subsection{Evaluation Benchmarks}
\label{sec:benchmarks}

\begin{table*}[!t]
\centering
\renewcommand{\arraystretch}{1.2}
\caption{Representative Video Understanding Benchmarks (Section~\ref{sec:benchmarks}). \textbf{Type}: \textbf{MCQ} (Multi-Choice), \textbf{OE} (Open-Ended), \textbf{Gen} (Generation), \textbf{Chat} (Dialogue). \textbf{Scale}: Number of QA pairs, videos, or annotations.}
\label{tab:benchmarks_main}
\scalebox{0.83}{
\begin{tabular}{l|c|c|l|c|c}
\toprule
\textbf{Benchmark} & \textbf{Year/Conf.} & \textbf{Source} & \textbf{Key Capabilities / Focus} & \textbf{Type} & \textbf{Scale} \\ \midrule
\multicolumn{6}{l}{\textit{\textbf{I. General Video Understanding}}} \\ \midrule
Video-MME~\cite{VideoMME} & CVPR 2024 & YouTube & Holistic perception across short/med/long durations & MCQ & 2.7K QA \\
MMBench-Video~\cite{MMBenchVideo} & NeurIPS 2024 & YouTube & Multi-shot perception; filtering static-solvable Qs & OE & 2K QA \\
Video-MME v2~\cite{VideoMMEv2} & arXiv 2026 & YouTube & Cohesive question groups; non-linear anti-guessing score & MCQ & 3.2K QA \\
MMWorld~\cite{MMWorld} & ICLR 2025 & 7 Disciplines & Multi-discipline causal \& real-world dynamic reasoning & MCQ & 6.6K QA \\ \midrule
\multicolumn{6}{l}{\textit{\textbf{II. Temporal \& Spatial Understanding}}} \\ \midrule
MVBench~\cite{MVBench} & CVPR 2024 & Public Sets & 20 fine-grained tasks (action, state, count, order) & MCQ & 4K QA \\
TempCompass~\cite{TempCompass} & ACL 2024 & Shutterstock & Temporal perception (speed, direction, attribute change) & MCQ & 7.5K QA \\
TOMATO~\cite{TOMATO} & ICLR 2025 & Diverse & Core multi-frame reasoning (rotation/direction/speed) & MCQ & 1.5K QA \\
E.T. Bench~\cite{liu2024etbench} & NeurIPS 2024 & Diverse & Event-level grounding, timestamp \& dense captioning & OE & 7.3K QA \\
TUNA~\cite{TUNA} & ACL 2025 & Diverse & Dense dynamic understanding (requires $>$16 frames) & MCQ/Gen & 2.4K QA \\
TimeLens~\cite{zhang2025timelens} & CVPR 2026 & VTG Sets & High-precision temporal grounding (re-annotated) & OE & 9.4K Annots \\
OMTG~\cite{xu2026omtg} & ICML 2026 & Diverse & First human labeling One-to-Many Temporal Grounding benchmark & OE & 340 Annots \\
TVGBench~\cite{TVGBench} & NeurIPS 2025 & Diverse & RL post-training surpasses SFT for temporal grounding & OE & -- \\
MotionBench~\cite{MotionBench} & CVPR 2025 & Hybrid & Fine-grained motion; camera vs. object motion & MCQ & 8K QA \\
DSI-Bench~\cite{DSIBench} & arXiv 2025 & Syn/Real & Dynamic spatial intelligence; observer motion & MCQ & 1.7K QA \\
STI-Bench~\cite{STIBench} & ICCV 2025 & Real & Quantitative spatio-temporal reasoning in 3D & MCQ & 2K QA \\
\multicolumn{6}{l}{\textit{\textbf{III. Complex Reasoning}}} \\ \midrule
V-STaR~\cite{VSTaR} & CVPR 2026 & VidSTG+ & Spatio-temporal reasoning (What-When-Where) & Gen & 2K Videos \\
MINERVA~\cite{MINERVA} & ICCV 2025 & YouTube & Counterfactual, causal, and goal-oriented reasoning & MCQ & 1.5K QA \\
VideoTT~\cite{VideoTT} & ICCV 2025 & Shorts & Truthfulness \& robustness against adversarial Qs & OE & 5K QA \\
MMR-V~\cite{MMRV} & arXiv 2025 & YouTube & Deep reasoning (implicit metaphors, irony, symbols) & MCQ & 1.2K QA \\
SEED-Bench-R1~\cite{SEEDBenchR1} & ACL 2026 & Ego/Ego4D & Reasoning and generalization in ego-centric views & MCQ & 50K QA \\
VideoReasonBench~\cite{VideoReasonBench} & ICLR 2026 & Syn/Real & Vision-centric latent-state \& counterfactual reasoning & OE & 1.4K QA \\
VideoZeroBench~\cite{VideoZeroBench} & arXiv 2026 & Web Long & Complex spatial event tracing in long videos & MCQ/OE & 500 QA \\ \midrule
\multicolumn{6}{l}{\textit{\textbf{IV. Long-Context \& Streaming Understanding}}} \\ \midrule
MLVU~\cite{MLVU} & CVPR 2025 & Movies+ & Holistic summary, plot analysis, needle retrieval & MCQ/OE & 3.1K QA \\
LongVideoBench~\cite{LongVideoBench} & ICCV 2025 & YouTube & Long-context referring reasoning \& relation & MCQ & 6.6K QA \\
LVBench~\cite{LVBench} & ICCV 2025 & Web & Extreme-length comprehension (avg. 68 min) & MCQ & 1.5K QA \\
ALLVB~\cite{ALLVB} & AAAI 2025 & Movies & Ultra-long context (avg. 114 min) comprehension & MCQ & 252K QA \\
CG-Bench~\cite{CGBench} & ICLR 2025 & Internet & Clue-grounded QA to prevent hallucination & MCQ/OE & 12K QA \\
StreamBench~\cite{StreamBench} & ICLR 2025 & Ego/Web & Real-time streaming understanding \& memory & Chat & 1.8K QA \\
OVO-Bench~\cite{OVOBench} & CVPR 2025 & Diverse & Online backward/real-time/forward behaviors & MCQ & 2.8K QA \\
SVBench~\cite{SVBench} & ICLR 2025 & Diverse & Temporal-jump streaming multi-turn dialogue & Chat & -- \\
OmniMMI~\cite{OmniMMI} & CVPR 2025 & 5 Open Sets & Proactive turn-taking \& anomaly alerting in streams & Chat & -- \\
Flash-VStream~\cite{zhang2025flash} & ICCV 2025 & Long Streams & Memory-based real-time long-stream evaluation & Chat & -- \\
RTV-Bench~\cite{RTVBench} & NeurIPS 2025 & EgoSchema & Continuous perception in real-time dynamic scenarios & MCQ & 4.6K QA \\ \midrule
\multicolumn{6}{l}{\textit{\textbf{V. Domain-Specific Knowledge}}} \\ \midrule
MMVU~\cite{MMVU} & CVPR 2025 & Expert & Expert-level reasoning (science, med, engineering) & MCQ & 3K QA \\
Video-MMMU~\cite{VideoMMMU} & arXiv 2025 & Lectures & Knowledge acquisition from professional videos & MCQ & 900 QA \\
ExpVid~\cite{ExpVid} & ICLR 2026 & JoVE & Scientific experiment understanding \& procedure & MCQ/Gen & 7.8K QA \\
Video-MMLU~\cite{VideoMMLU} & ICCV 2025 & Lectures & STEM lecture understanding (Theorem/Problem solving) & OE & 15.7K QA \\
BEAR~\cite{BEAR} & arXiv 2025 & Embodied & Atomic embodied capabilities (pointing to planning) & MCQ/OE & -- \\ \midrule
\multicolumn{6}{l}{\textit{\textbf{VI. Omnimodal Collaboration}}} \\ \midrule
WorldSense~\cite{WorldSense} & ICLR 2026 & Audio-Vis & Strict audio-visual synergy (speech, music, env) & MCQ & 3.1K QA \\
OmniVideoBench~\cite{OmniVideoBench} & ICLR 2026 & Web & Audio-visual reasoning with explicit CoT & MCQ/OE & 1K QA \\
LongVALE~\cite{LongVALE} & CVPR 2025 & Long Vids & Vision-audio-language-event dense alignment & Gen/OE & 105K Events \\
LongInsightBench~\cite{LongInsightBench} & arXiv 2025 & FineVideo & Long omnimodal; intra/inter-event reasoning (avg 9 min) & MCQ & 4.8K QA \\
LVOmniBench~\cite{LVOmniBench} & arXiv 2026 & Web & Ultra-long audio-video (avg. 34.5 min) comprehension & MCQ & -- \\
MMOU~\cite{MMOU} & arXiv 2026 & Web & Mandatory multi-modal; vision/text-only fail & MCQ/OE & 15K QA \\
Omni-Captioner~\cite{OmniCaptioner} & ICLR 2026 & Diverse & Fine-grained omni perception via cloze protocol & Gen & 69.6K Cloze \\
LiViBench~\cite{LiViBench} & AAAI 2026 & Livestream & Interactive livestream understanding \& culture & MCQ & -- \\ \bottomrule
\end{tabular}%
}
\end{table*}

As Video MLLMs evolve from simple perception to complex cognition, the evaluation landscape has shifted from traditional metrics to comprehensive benchmarks that assess a broader range of cognitive capabilities. 
We categorize existing benchmarks into six dimensions: General Video Understanding, Temporal and Spatial Understanding, Complex Reasoning, Long-Context and Streaming Understanding, Domain-Specific Knowledge, and Omnimodal Understanding. 
A comprehensive overview of representative benchmarks is provided in Table~\ref{tab:benchmarks_main}.

\noindent\textbf{General Video Understanding.}
Benchmarks in this category are designed to evaluate holistic perception across diverse domains and durations. Video-MME~\cite{VideoMME} constructs a comprehensive dataset covering short, medium, and long videos to assess capability across different temporal scales, and its successor Video-MME v2~\cite{VideoMMEv2} further strengthens evaluation rigor by introducing cohesive question-group designs with non-linear scoring that penalizes blind guessing, while sourcing all videos from late 2025 to prevent pre-training data leakage. MMBench-Video~\cite{MMBenchVideo} incorporates a rigorous ``video-exclusivity'' filtering mechanism, using GPT-4V to remove questions that can be answered by a single static frame, thereby ensuring the evaluation focuses on temporal dynamics rather than static recognition. MVBench~\cite{MVBench} further defines a systematic taxonomy of temporal tasks from action sequencing to counterfactual inference across diverse visual contexts. Extending evaluation into multi-discipline real-world scenarios, MMWorld~\cite{MMWorld} spans seven academic disciplines and tests causal and domain-specific reasoning over real-world video dynamics.

\noindent\textbf{Temporal and Spatial Understanding.}
This domain focuses on the dynamic nature of video through fine-grained tasks related to temporal perception, motion analysis, and spatial reasoning. On the temporal side, TempCompass~\cite{TempCompass} defines a targeted taxonomy for temporal attributes---action, speed, direction, and attribute change---and constructs ``conflict videos'' to verify that models are not exploiting static biases, while TOMATO~\cite{TOMATO} further isolates core multi-frame temporal reasoning such as rotation, direction, and speed that cannot be resolved by common sense or a single frame. To address the need for higher precision, TimeLens~\cite{zhang2025timelens} provides recalibrated high-precision temporal annotations for grounding evaluation and demonstrates the effectiveness of reinforcement learning for temporal localization, and TUNA~\cite{TUNA} specifically filters for questions requiring at least sixteen frames of context, ensuring that evaluations capture genuinely fine-grained holistic dynamics rather than sparse keyframe shortcuts. Beyond temporal grounding, E.T. Bench~\cite{liu2024etbench} assesses event-level open-ended understanding including fine-grained retrieval, timestamp prediction, and dense captioning, revealing that most MLLMs struggle to output structured temporal references. TVGBench~\cite{TVGBench} and TimeScope~\cite{TimeScope} push the boundaries of temporal grounding further: TVGBench demonstrates that reinforcement learning post-training with minimal data can surpass full supervised fine-tuning for temporal localization, while TimeScope targets task-oriented grounding in long videos where traditional methods suffer steep performance drops.

For spatial and motion understanding, MotionBench~\cite{MotionBench} and DSI-Bench~\cite{DSIBench} systematically evaluate the ability to distinguish between camera motion and object motion, with DSI-Bench specifically probing observer-scene and observer-object spatial relationships. STI-Bench~\cite{STIBench} goes a step further by assessing quantitative 3D spatio-temporal reasoning grounded in real-world scenarios such as autonomous driving and indoor reconstruction, while SI-Bench~\cite{SIBench} consolidates nearly twenty spatial reasoning datasets to comprehensively test visual spatial intelligence including environment navigation and embodied planning. SVAG-Bench~\cite{SVAGBench} extends traditional grounding by introducing a multi-instance spatio-temporal setting that requires simultaneous tracking and localization of multiple objects, with a novel joint evaluation metric.

\noindent\textbf{Complex Reasoning.}
Beyond perception, ``System 2'' benchmarks assess the depth of cognitive processing. VCR-Bench~\cite{VCRBench} and V-STaR~\cite{VSTaR} emphasize the \emph{process} of reasoning, requiring explicit Chain-of-Thought (CoT) traces or structured ``What-When-Where'' outputs to verify the logic behind answers rather than merely the final result. VideoReasonBench~\cite{VideoReasonBench} takes a vision-centric approach, using programmatically synthesized videos to test fine-grained perception, latent state tracking, and counterfactual prediction, where even the strongest reasoning models show severely limited performance. SEED-Bench-R1~\cite{SEEDBenchR1} targets next-action prediction in egocentric daily scenarios, testing models' ability to reason about procedural planning from first-person perspectives. Know-Show~\cite{KnowShow} further raises the bar by requiring models to not only reason correctly but also ground their answers by localizing supporting spatio-temporal evidence in the video, bridging the gap between reasoning accuracy and visual accountability.

In terms of logic and robustness, MINERVA~\cite{MINERVA} requires models to combine multiple reasoning skills per question---temporal, numerical, and counterfactual---with human-annotated reasoning traces for interpretable evaluation, while MMR-V~\cite{MMRV} challenges implicit reasoning over non-literal content such as irony, metaphor, and counter-intuitive narratives through extended multi-option questions. VideoTT~\cite{VideoTT} employs adversarial questioning strategies with deliberately misleading prompts to evaluate model truthfulness and robustness against deceptive cues. VideoZeroBench~\cite{VideoZeroBench} pushes further by targeting complex spatial event tracing in long videos, revealing that even frontier models achieve extremely low accuracy without explicit spatio-temporal cropping assistance.

\noindent\textbf{Long-Context and Streaming Understanding.}
Addressing the challenge of extended durations, MLVU~\cite{MLVU} compiles videos ranging from minutes to two hours with tasks spanning detail retrieval to global topic reasoning, while LongVideoBench~\cite{LongVideoBench} specifically targets long-context referential reasoning through interleaved video-language understanding. ALLVB~\cite{ALLVB} scales to feature-film-length videos with an average duration exceeding one hundred minutes and over a quarter million QA pairs, testing needle-in-a-haystack retrieval, emotion recognition, and event detection at the movie level. LVBench~\cite{LVBench} further extends to extreme lengths with an average duration of nearly seventy minutes, while AdaVideoRAG~\cite{xue2025adavideorag} introduces a retrieval-augmented evaluation framework for ultra-long videos spanning up to nearly two hours, probing fact extraction, cross-segment causal reasoning, and external knowledge integration. CG-Bench~\cite{CGBench} introduces a clue-grounding mechanism that requires models to identify specific video intervals that support their answers, making evaluation more faithful and interpretable.

In streaming and real-time scenarios, StreamBench~\cite{StreamBench} simulates continuous video inputs and multi-round interactions over unfolding timelines, while OVO-Bench~\cite{OVOBench} models three key online behaviors: backward tracing, real-time perception, and forward anticipation. SVBench~\cite{SVBench} introduces temporal jump evaluation that forces models to handle cross-segment temporal dependencies in streaming contexts, and RTV-Bench~\cite{RTVBench} uniquely designs questions whose correct answers change as the video progresses, testing dynamic continuous perception rather than static one-shot comprehension. MT-Video-Bench~\cite{MTVideoBench} evaluates holistic multi-turn video dialogue ability across successive conversational rounds including cross-scene reasoning and proactive interaction. OmniMMI~\cite{OmniMMI} additionally evaluates proactive capabilities such as autonomous turn-taking and anomaly alerting in streaming contexts, revealing that current models have virtually no capacity for proactive interaction. Flash-VStream~\cite{zhang2025flash} proposes a memory-based evaluation protocol for real-time understanding of extremely long video streams under tight memory and latency constraints.

\noindent\textbf{Domain-Specific Knowledge.}
Benchmarks in this category evaluate the ability to combine visual perception with specialized expertise. MMVU~\cite{MMVU} and Video-MMMU~\cite{VideoMMMU} both draw from professional content---scientific, medical, engineering, and humanities---but adopt complementary evaluation philosophies: MMVU focuses on expert-level knowledge-intensive reasoning, while Video-MMMU uniquely measures knowledge \emph{acquisition} by quantifying how much a model can learn from instructional videos rather than simply recalling pre-trained knowledge. Video-MMLU~\cite{VideoMMLU} extends multi-discipline lecture understanding to cover a broader range of STEM subjects with a substantially larger question set. ExpVid~\cite{ExpVid} narrows the focus to scientific experiment videos, assessing a three-stage cognitive pipeline from perception through procedural comprehension to scientific reasoning, using content sourced from peer-reviewed video journals. BEAR~\cite{BEAR} shifts to embodied intelligence, evaluating atomic capabilities from low-level pointing and trajectory understanding to high-level planning, providing a fine-grained diagnostic of embodied perception and interaction readiness.

\noindent\textbf{Omnimodal Understanding.}
Validating true multimodal fusion, benchmarks in this category enforce strict audio-visual dependencies to ensure that questions cannot be answered by a single modality. WorldSense~\cite{WorldSense} curates questions requiring genuine audio-visual synergy across speech, environmental sounds, and music modalities, while OmniVideoBench~\cite{OmniVideoBench} implements a rigorous multi-stage purification pipeline to verify that every question demands cross-modal reasoning, covering thirteen fine-grained task categories. LongVALE~\cite{LongVALE} extends omnimodal evaluation to dense event-level annotation over long videos with vision-audio-language-event alignment, providing over one hundred thousand event annotations across thousands of videos.

Scaling omnimodal evaluation to longer durations, LongInsightBench~\cite{LongInsightBench} focuses on long videos averaging nine minutes with multi-model collaborative annotation to assess both intra-event local reasoning and inter-event long-range reasoning, while LVOmniBench~\cite{LVOmniBench} pushes to an average of over thirty minutes, revealing that open-source models largely fail at such extended audio-visual comprehension. MMOU~\cite{MMOU} provides a massive multi-task benchmark with over fifteen thousand QA pairs where every question provably requires multiple modalities, confirming that neither vision-only nor text-only models can succeed. Omni-Captioner~\cite{OmniCaptioner} evaluates fine-grained omnimodal detailed perception through a novel cloze-style protocol, testing whether models can capture and articulate subtle audio-visual details. At the adversarial frontier, OMD-Bench~\cite{OMDBench} deliberately introduces cross-modal information conflicts---mismatched visual, auditory, and textual signals---to probe modality robustness and calibrated abstention, revealing severe overconfidence under corrupted multimodal inputs. LiViBench~\cite{LiViBench} uniquely targets interactive livestream video understanding, covering domain-specific cultural elements such as live-streaming interactions and gift-giving that demand specialized omnimodal comprehension.
\section{Future Directions}
\label{sec:future_direction}

\subsection{Spatial Reasoning in Video Understanding}
Spatial reasoning at both the object and scene levels remains a crucial frontier for LLM-based video understanding. 
Current video LLMs often excel at holistic scene description but struggle with fine-grained spatial details, for example, precisely localizing or tracking specific objects and their relationships over time~\cite{yang2025visual,zhang2025videorefer}.
Conversely, building a coherent global model of a scene (like the 3D layout of an environment) from video is inherently difficult due to limited viewpoints, occlusions, and the need to maintain spatial consistency across frames~\cite{zhang2025spatial}. 
Therefore, bridging this object-level and scene-level gap is essential for advanced applications such as detailed video question answering, robotic perception, and embodied navigation, which require an understanding of where things are in the scene and how they relate in space. 
Achieving human-like spatial understanding will require overcoming current limitations, such as poor long-term object tracking and shallow geometric comprehension in current models.

Emerging research directions are beginning to tackle these challenges. At the object level, new Video~LLM architectures integrate dedicated visual encoders to improve fine-grained spatial perception~\cite{zhang2025videorefer}.
At the scene level, some approaches~\cite{qi2025gpt4scene} introduce explicit spatial representations to fuse multi-view cues and capture global layout for reasoning.
In addition, researchers are exploring structured reasoning techniques to guide spatial understanding~\cite{zhang2025spatial}.
For example, using chain-of-thought prompting or step-by-step query decomposition to help models infer spatial relations and geometry in video without modifying the underlying architecture. 
%
%
Progress in this direction, through better spatial memory mechanisms, multimodal world models, and spatially grounded training paradigms, can enable video LLMs to reliably reason about physical environments, powering applications ranging from long-horizon video analysis to autonomous robot planning in dynamic scenes.

\subsection{Multi-Video and Multi-Segment Temporal Grounding}
Real video applications rarely involve a single, clean, continuous clip. 
Users often watch or create highlight compilations, reaction videos, or collections of related videos. 
In this setting, temporal grounding goes beyond locating frames. 
The model must understand the content across segments and identify evidence that truly matches the user's intent, rather than simply predicting timestamps for a single salient moment.

Although video temporal grounding has progressed rapidly, most methods still assume one video as input. Time-aware tuning and improved time tokenization make timestamp prediction more natural, but they do not directly solve cross-segment ambiguity~\cite{ren2023timechat,huang2024vtimellm}. 
Efficiency methods improve coverage under a limited context, yet they typically search within a single timeline~\cite{yu2023sevila}. 
Structured decoding reduces underspecified outputs, but repeated patterns and replays across edits still cause boundary errors~\cite{guo2025trace,li2025unitime,guo2025tartvg}. Multi-segment grounding is starting to be studied, but multi-video grounding remains far from solved.
A useful direction is to model multi-video grounding as \emph{set-based retrieval + refinement}. 
A simple pipeline for hierarchical grounding: (1) retrieve candidate segments across videos; (2) refine start/end times inside each segment. Edit-aware cues can further help, such as predicting cut points or segment IDs and using them as anchors. 
Finally, verifiable post-training with IoU-style rewards can encourage accurate and stable boundaries under large search spaces~\cite{wang2025time,zhang2025timelens}.

\subsection{Hour-Scale Video Understanding with Structured Memory}

Moving from minutes to hours changes the problem. Many tasks (meetings, lectures, sports, daily-life streams) need \emph{second-level details} and also \emph{long-range dependencies}. A model must capture rare but decisive moments, track entities over long periods, and connect evidence that may be far apart in time. This requires stronger memory, not just longer context.

Current approaches mainly use compression, sparse selection, or periodic summaries. These methods reduce costs, but they often lose key details or break long-range dependencies~\cite{song2024moviechat,shen2024longvu,lan2024vidcompress,diko2025rewind}. Event-based and hierarchical memory improves scalability, but it raises practical issues: when to write, what to update, and how to avoid summary drift~\cite{cheng2025enhancing,santos2025video,lin2025unleashing,azad2025hierarq,faure2025hermes}. Agentic systems add external memory and retrieval, but they can be expensive and may fail when retrieval is slightly wrong~\cite{fan2024videoagent,xue2025adavideorag,zuo2025videolucy}.

A promising direction is \emph{structured multi-level memory} with \emph{evidence pointers}. One practical design is three tiers: a short buffer for recent fine-grained evidence, an event memory that stores temporally bounded episodes, and a long-term store for entities and relations. Memory writing and forgetting should be learned, so the model keeps rare but important events and drops redundant content. Retrieval should return both a short summary and the supporting time spans, so the model can recheck evidence when needed. Streaming-style forgetting is also critical for hour-scale inputs~\cite{lin2025unleashing,yang2025streammem,chen2025streamingtom}.

\subsection{Efficient and Verifiable Video Reasoning}

Long-video reasoning must balance \emph{cost} and \emph{faithfulness}. It is too expensive to process all frames, but it is also risky to reason without checking evidence. This motivates efficient and verifiable reasoning: the model should selectively inspect the video and present explicit evidence (e.g., timestamps, key frames, boxes) that can be checked.

We already have strong components. Efficient watching reduces redundant inputs via frame selection, token compression, and cache optimization~\cite{tang2025aks,zhang2025qframe,fu2025framefusion,qin2025videoxl2,wang2025adaretake}. Thinking-with-videos methods reduce hallucination by rechecking evidence during reasoning, either via tool use or via structured outputs~\cite{zhang2025thinking,yan2025videochat,zhang2025rewatch,wang2025pixel,ouyang2025conan,meng2025open,wang2025video-thinker}. However, many systems still inspect too much, repeat similar queries, or output evidence that looks plausible but is not minimal. Training often optimizes answers more than evidence quality.

A useful direction is to treat grounded reasoning as \emph{budgeted evidence search}. Training can jointly optimize: answer correctness, evidence alignment (temporal/spatial IoU), and evidence compactness. This can be done with verifiable RL or verifier-guided preference optimization~\cite{wang2025time,zhang2025timelens,li2025veripo,huang2025vistadpo}. Another direction is uncertainty-aware inspection: the model requests additional evidence only when its current reasoning is uncertain. Finally, standard structured schemas for evidence (timestamps, boxes, grounded captions) can make training and evaluation more consistent across tasks~\cite{meng2025open,gu2025thinking,li2025unitime}.

\subsection{Streaming Egocentric Video Understanding}

Streaming egocentric video is different from offline third-person benchmarks. The stream is long and continuous, viewpoints change quickly, and interactions are frequent. The model must update its state online, manage memory under latency constraints, and keep a coherent view of the user's goals and the environment. This setting is important for proactive assistants and embodied agents, where timing matters (e.g., when to speak or intervene).
Recent work has improved fine-grained grounding and interaction reasoning in egocentric videos, and has explored reinforcement learning for 4D world modeling and navigation~\cite{liang2025fine,wu2025st,qi2025vln}. Ultra-long egocentric reasoning further motivates hierarchical retrieval and tool coordination~\cite{tian2025ego}. Proactive timing is also emerging as a key capability in streaming settings~\cite{zhang2025eyes,wang2025egosocialbenchmarkingproactiveintervention}. 
On the systems side, streaming memory methods keep constant memory via pruning, compression, and hierarchical storage, but they are not yet tightly coupled with interaction goals~\cite{yang2025streammem,kim2025infinipotv,chen2025streamingtom,xiong2025streaming,zhang2024flash,ning2025livevlm}.

Future work should focus on \emph{stateful, goal-driven streaming memory}. One direction is to keep an explicit task state that controls what is stored and what is ignored. Another is event-triggered writing: store interaction episodes as structured records rather than uniform summaries. Proactive retrieval can further bring back relevant past evidence before it is needed (e.g., where an object was last seen). Finally, evaluation should go beyond static QA and include timing, stability under updates, and safe intervention, which can be optimized in embodied settings~\cite{tian2025ego,zhang2025eyes,qi2025vln}.


\section{Conclusion}
\label{sec:conclusion}

This survey reviews MLLM-based video understanding through the human view: \textit{watching}, \textit{remembering}, and \textit{reasoning}.
We summarize progress in spatio-temporal perception, efficient observation, memory construction and retrieval, and reasoning-centric training and evaluation.
Recent work shows a clear shift from simple input compression and answer generation toward structured memory, streaming systems, and explicit evidence-grounded reasoning.
These trends highlight the importance of scalable, memory-aware, and verifiable video intelligence.


\ifCLASSOPTIONcaptionsoff
  \newpage
\fi



{
\bibliographystyle{IEEEtran}
\bibliography{IEEEabrv,
    main
}
}

%








\end{document}